\documentclass[pageno]{jpaper}
\usepackage{mathptmx} 
\usepackage{balance}
\usepackage{authblk}
\usepackage{booktabs}
\usepackage{todonotes}
\usepackage{epsfig}
\usepackage{graphicx}
\usepackage{amsmath}
\usepackage{amssymb}
\usepackage{lipsum}
\usepackage{subfig}
\usepackage{color}
\usepackage{cite}
\usepackage{url}

\usepackage[normalem]{ulem}
\usepackage{bigstrut}
\usepackage{multirow}
\usepackage{siunitx}
\usepackage{ulem}
\usepackage{listings}
\usepackage{eso-pic}
\usepackage{tikz}
\usepackage[colorlinks  = true,
            linkcolor   = blue,
            citecolor   = red,
            urlcolor    = blue,
            anchorcolor = blue
            breaklinks,
            pagebackref,
            draft=false]{hyperref}   

\newcommand*\circled[2]{\tikz[baseline=(char.base)]{
            \node[shape=circle,fill=black,inner sep=1pt] (char) {\textcolor{#1}{{\footnotesize #2}}};}}

\ifx\figurename\undefined \def\figurename{Figure}\fi
\renewcommand{\figurename}{Fig.}
\renewcommand{\paragraph}[1]{\textbf{#1}~~}

\newcommand{\Sect}[1]{Sec.~\ref{#1}}
\newcommand{\Fig}[1]{Fig.~\ref{#1}}
\newcommand{\Tbl}[1]{Table~\ref{#1}}
\newcommand{\Equ}[1]{Equ.~\ref{#1}}

\newcommand{\specialcell}[2][c]{\begin{tabular}[#1]{@{}l@{}}#2\end{tabular}}

\newcommand{\RNum}[1]{\uppercase\expandafter{\romannumeral #1\relax}}

\graphicspath{{figs/}}

\usepackage{authblk}

\title{Euphrates: Algorithm-SoC Co-Design for Low-Power Mobile Continuous Vision}
\author[1]{Yuhao Zhu}
\author[2]{\quad\quad Anand Samajdar}
\author[3]{\quad\quad Matthew Mattina}
\author[3]{\quad\quad Paul Whatmough}
\affil[1]{University of Rochester}
\affil[2]{Georgia Institute of Technology}
\affil[3]{ARM Research}

\date{}

\begin{document}
\maketitle


\begin{abstract}
Continuous computer vision (CV) tasks increasingly rely on convolutional neural
networks (CNN). However, CNNs have massive compute demands that far exceed the
performance and energy constraints of mobile devices. In this paper, we propose and develop an algorithm-architecture co-designed system, Euphrates, that simultaneously improves
the energy-efficiency and performance of continuous vision tasks.


Our key observation is that changes in pixel data between consecutive frames
represents visual motion. We first propose an algorithm that leverages this motion
information to relax the number of expensive CNN inferences required by
continuous vision applications.
We co-design a mobile System-on-a-Chip (SoC) architecture to maximize the
efficiency of the new algorithm. The key to our architectural augmentation is to \textit{co-optimize different SoC IP blocks in the vision pipeline collectively}. Specifically, we propose to expose the motion data that is naturally generated by the Image Signal Processor (ISP) early in the vision pipeline to the CNN engine.
Measurement and synthesis results show that Euphrates achieves up to 66\%
SoC-level energy savings (4$\times$ for the vision computations), with only 1\%
accuracy loss.
\end{abstract}


\section{Introduction}
\label{sec:intro}

Computer vision (CV) is the cornerstone of many emerging application domains,
such as advanced driver-assistance systems (ADAS) and augmented reality (AR).
Traditionally, CV algorithms were dominated by hand-crafted features (e.g.,
Haar~\cite{violajones} and HOG~\cite{hog}), coupled with a classifier such as a
support vector machine (SVM)~\cite{svm}. 
These algorithms have low complexity and are practical in constrained
environments, but only achieve moderate accuracy.
Recently, convolutional neural networks (CNNs) have rapidly displaced
hand-crafted feature extraction, demonstrating
significantly higher accuracy on a range of CV tasks including image classification~\cite{vgg}, object
detection~\cite{fasterrcnn, ssd, yolo}, and visual tracking~\cite{mdnet, eco}.

\begin{figure}[t]
\vspace{-2pt}
  \centering
  \includegraphics[trim=0 0 0 0, clip, width=.9\columnwidth]{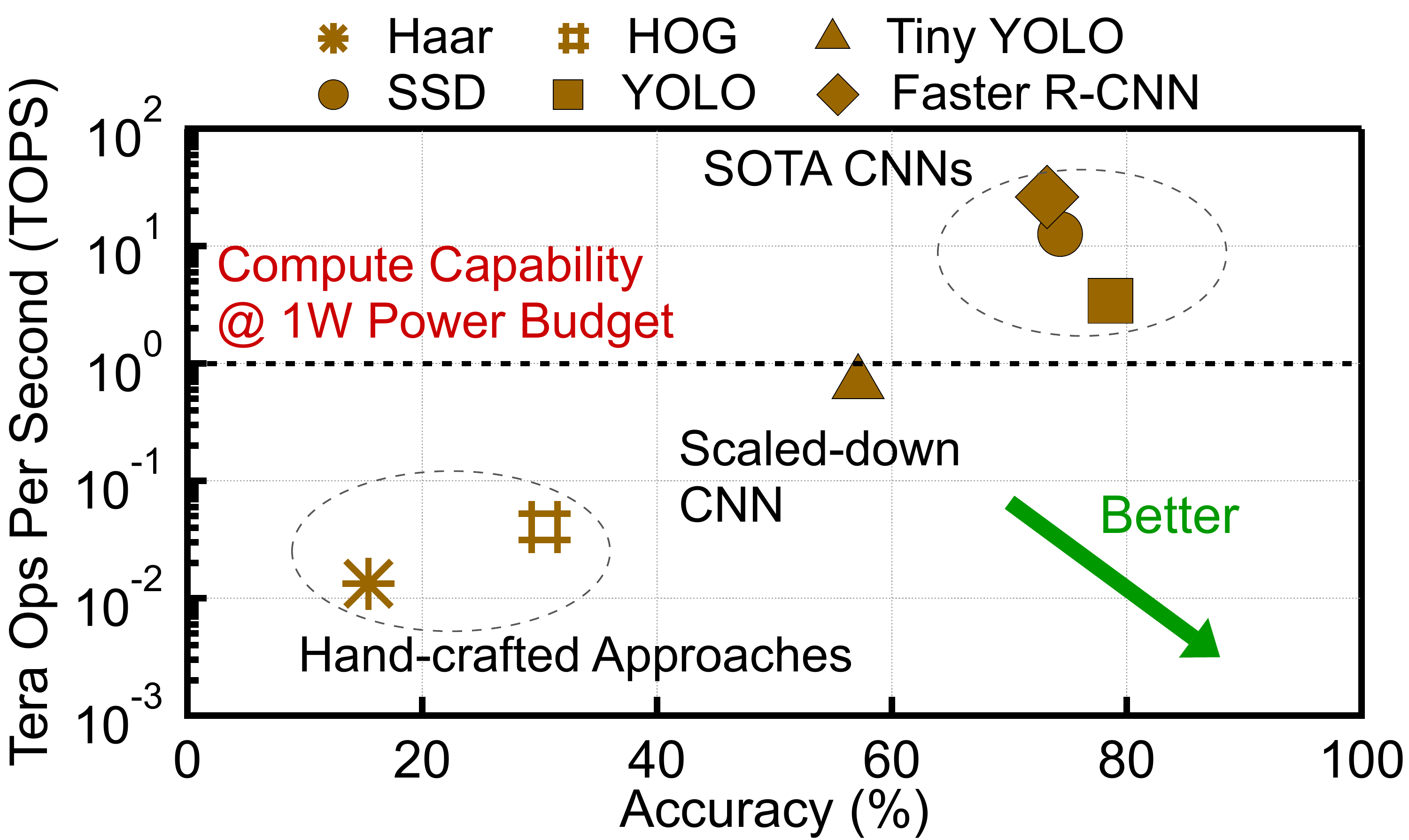}
  \caption{\small{Accuracy and compute requirement (TOPS) comparison between
  object detection techniques. Accuracies are measured against the widely-used
  PASCAL VOC 2007 dataset~\cite{voc2007}, and TOPS is based on the 480p (640$\times$480) resolution common in smartphone cameras.}}
  \label{fig:gap}
\end{figure}

This paper focuses on \textit{continuous vision} applications that
extract high-level semantic information from \textit{real-time} video streams. Continuous
vision is challenging for mobile architects due to its enormous compute
requirement~\cite{mobilemlatarm}. Using object
detection as an example, \Fig{fig:gap} shows the compute requirements measured
in Tera Operations Per Second (TOPS) as well as accuracies between different detectors
under 60 frames per second (FPS). As a reference, we also overlay the 1 TOPS line, which represents
the peak compute capability that today's CNN accelerators offer under a
typical 1~W mobile power budget~\cite{myriad2, myriadxbrief}. We find that today's
CNN-based approaches such as YOLOv2~\cite{yolov2}, SSD~\cite{ssd}, and Faster
R-CNN~\cite{fasterrcnn} all have at least one order of magnitude higher compute
requirements than accommodated in a mobile device. Reducing the CNN complexity (e.g.,
Tiny YOLO~\cite{yolo}, which is a heavily truncated version of YOLO with 9/22 of its
layers) or falling back to traditional hand-crafted features such as
Haar~\cite{acf} and HOG~\cite{fastestdpm} lowers the compute demand, which, however, comes at a
significant accuracy penalty.



The goal of our work is to 
improve the compute efficiency of continuous vision with small accuracy loss, thereby enabling
new mobile use cases. The key idea is to exploit the motion information
inherent in real-time videos. Specifically, today's continuous vision
algorithms treat each frame as a standalone entity and thus execute an entire
CNN inference on every frame. However, pixel changes across consecutive frames
are not arbitrary; instead, they represent visual object motion. We propose a
new algorithm that leverages the temporal pixel motion to synthesize vision
results with little computation while avoiding expensive CNN inferences on many
frames.

Our main architectural contribution in this paper is to co-design the mobile SoC architecture to support the new algorithm. Our SoC
augmentations harness two insights. First, we can greatly improve the compute efficiency while simplifying the architecture design by \textit{exploiting the synergy between
different SoC IP blocks}. Specifically, we observe that the pixel motion
information is naturally generated by the ISP early in
the vision pipeline owing to ISP's inherent algorithms, and thus can
be obtained with little compute overhead. We augment the SoC with a lightweight
hardware extension that exposes the motion information to the vision engine. In contrast, prior work extracts motion information manually, either offline from an already compressed video~\cite{mvcnn, mvar}, which does not apply to real-time video streams, or by calculating the motion information at runtime at a performance cost~\cite{inoue2016adaptive, fastyolo}.

Second, although the new algorithm is light in compute, implementing it in
software is energy-inefficient from a system perspective because it
would frequently wake up the CPU.
We argue that always-on continuous computer vision should be task-autonomous, i.e., free from interrupting the CPU. Hence, we introduce the concept of a \textit{motion controller}, which is a new hardware IP that autonomously sequences the vision
pipeline and performs motion extrapolation\textemdash all without interrupting
the CPU. The motion controller's 
microarchitecture resembles a micro-controller, and thus incurs very low design cost.



We develop Euphrates, a proof-of-concept system of our algorithm-SoC co-designed
approach. We evaluate Euphrates on two tasks, object
tracking and object detection, that are critical to many
continuous vision scenarios such as ADAS and AR. Based on real hardware measurements and RTL
implementations, we show that Euphrates doubles the object detection rate while
reducing the SoC energy by 66\% at the cost of less than 1\% accuracy loss; it also
achieves 21\% SoC energy saving at about 1\% accuracy loss for object tracking.

In summary, we make the following contributions:

\begin{itemize}
  \item To our knowledge, we are the first to exploit sharing motion data across the ISP and other IPs in an SoC.
  \item We propose the Motion Controller, a new IP that autonomously coordinates
  the vision pipeline during CV tasks, enabling ``always-on'' vision with very
  low CPU load.
  \item We model a commercial mobile SoC, validated with hardware measurements and RTL implementations, and achieve significant energy savings and frame rate improvement.
\end{itemize}

The remainder of the paper is organized as follows. \Sect{sec:background} introduces the background. \Sect{sec:algo} and \Sect{sec:arch} describe the motion-based algorithm and the co-designed architecture, respectively. \Sect{sec:exp} describes the evaluation methodology, and \Sect{sec:eval} quantifies the benefits of Euphrates. \Sect{sec:disc} discusses limitations and future developments. \Sect{sec:related} puts Euphrates in the context of related work, and \Sect{sec:conc} concludes the paper.

\section{Background and Motivation}
\label{sec:background}

We first give an overview of the continuous vision
pipeline from the software and hardware perspectives~(\Sect{sec:background:pipe}). In particular, we highlight an important
design trend in the vision frontend where ISPs are increasingly incorporating motion estimation, which
we exploit in this paper~(\Sect{sec:background:isp}). Finally, we briefly
describe the block-based motion estimation algorithm and its data structures that are used in this paper~(\Sect{sec:background:mv}).

\begin{figure}[t]
  \centering
  \includegraphics[trim=0 0 0 0, clip, width=\columnwidth]{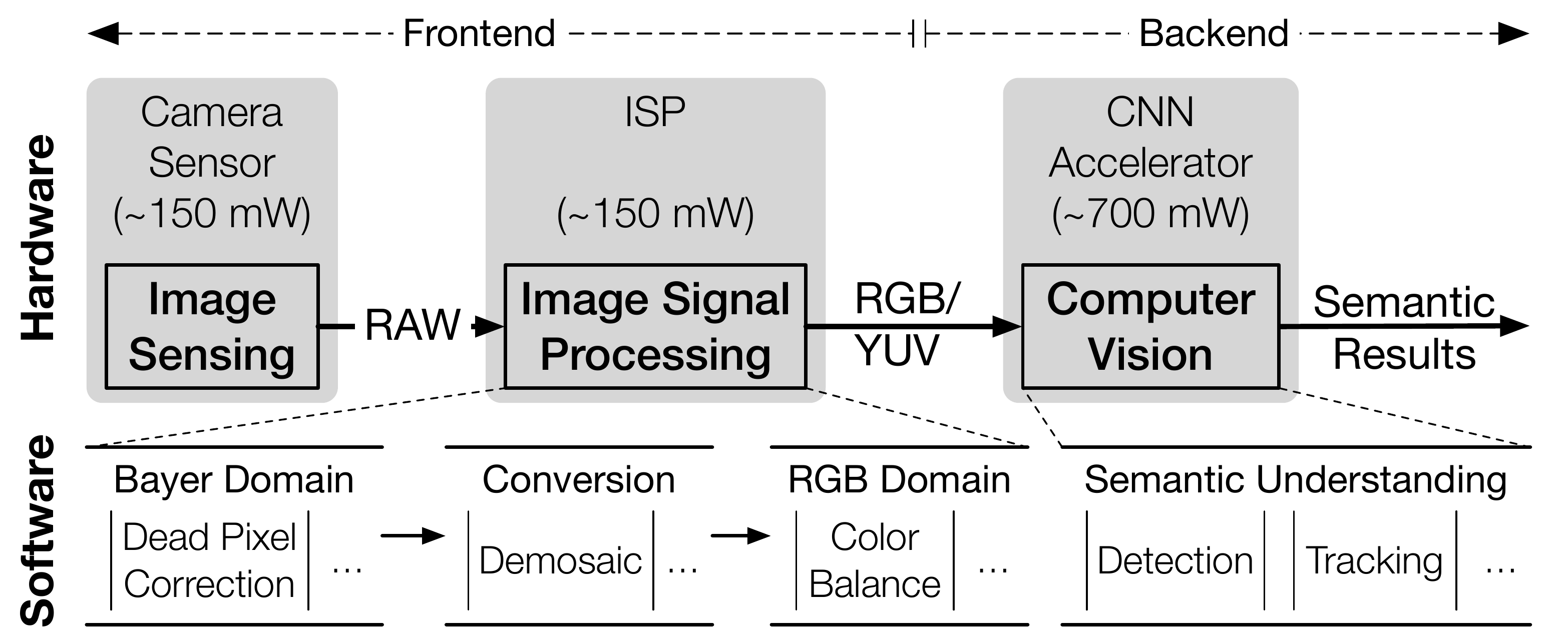}
  \caption{\small{A typical continuous computer vision pipeline.}}
  \label{fig:pipe}
\end{figure}

\subsection{The Mobile Continuous Vision Pipeline}
\label{sec:background:pipe}

The continuous vision pipeline consists of two parts: a frontend and a backend,
as shown in \Fig{fig:pipe}. The frontend prepares pixel data for the backend,
which in turn extracts semantic information for high-level decision making.

The frontend uses (off-chip) camera sensors to capture light and produce
RAW pixels that are transmitted to the mobile SoC, typically over the MIPI
camera serial interface (CSI)~\cite{mipicsi2}. Once on-chip, the Image Signal
Processor (ISP) transforms the RAW data in the Bayer domain to pixels in the
RGB/YUV domain through a series of algorithms such as dead pixel correction,
demosacing, and white-balancing.
In architecture terms, the ISP is a specialized IP block in a mobile SoC,
organized as a pipeline of mostly stencil operations on a set of local SRAMs
(``line-buffers''). The vision frontend typically stores frames in the main memory
for communicating with the vision backend due to the large size of the image data.



The continuous vision backend extracts useful information from images through semantic-level tasks such as object detection.
Traditionally, these algorithms are
spread across DSP, GPU, and CPU. Recently,  the rising compute intensity of CNN-based algorithms and the pressing need for energy-efficiency have urged mobile SoC vendors to deploy
dedicated CNN accelerators. Examples include the Neural
Engine in the iPhoneX~\cite{applexaicore} and the CNN co-processor in the HPU~\cite{hpuaicore}.

\paragraph{Task Autonomy} During continuous vision tasks, different SoC components autonomously coordinate among each other with minimal CPU intervention, similar to during phone
calls or music playback~\cite{vip}. Such a task autonomy frees the CPU to
either run other OS tasks to maintain system responsiveness, or stay in the stand-by mode to save power. As a comparison, the typical power
consumption of an image sensor, an ISP, and a CNN accelerator combined is about
1~W (refer to~\Sect{sec:exp:hw} for more details), whereas the CPU cluster alone
can easily consume over 3~W~\cite{googleglasschar, mobilecpu}. Thus, we must maintain task autonomy by
minimizing CPU interactions when optimizing energy-efficiency for continuous vision
applications.

\paragraph{Object Tracking and Detection} This paper focuses on two continuous
vision tasks, object tracking and detection, as they are key enablers for emerging mobile application domains such as AR\cite{qualcommobjdet} and ADAS\cite{nvadas}. Such CV tasks are also prioritized by commercial IP vendors, such as ARM, who recently announced standalone Object Detection/Tracking IP~\cite{armod}.

Object tracking involves localizing a moving object across frames by predicting the coordinates of its bounding box, also known as
a region of interest (ROI). Object detection refers to simultaneous object classification
and localization, usually for multiple objects. Both detection and
tracking are dominated by CNN-based techniques. For instance, the
state-of-the-art object detection network YOLO~\cite{yolo} achieves 37\% higher
accuracy than the best non-CNN based algorithm DPM~\cite{dpm}. Similarly,
CNN-based tracking algorithms such as MDNet~\cite{mdnet} have shown over 20\%
higher accuracy compared to classic hand-crafted approaches such as
KCF~\cite{kcf}.


\subsection{Motion Estimation in ISPs}
\label{sec:background:isp}

Perceived imaging quality has become a strong product
differentiator for mobile devices.
As such, ISPs are starting to integrate sophisticated computational photography algorithms
that are traditionally performed as separate image enhancement tasks, possibly
off-line, using CPUs or GPUs. A classic example is recovering high-dynamic range
(HDR)~\cite{hdr}, which used to be implemented in software (e.g., in Adobe
Photoshop~\cite{pshdr}) but is now directly built into many consumer camera
ISPs~\cite{malicamera, snapdragon835, denalimcisp, fpgaasicisp}.



Among new algorithms that ISPs are integrating is \textit{motion estimation}, which
estimates how pixels move between consecutive frames. Motion estimation is at
the center of many imaging algorithms such as temporal denoising, video
stabilization (i.e., anti-shake), and frame upsampling. For instance, a
temporal denoising algorithm~\cite{liu2010high, ji2010robust} uses pixel motion
information to replace noisy pixels with their noise-free counterparts in the
previous frame. Similarly, frame upsampling~\cite{choi2007motion} can
artificially increase the frame rate by interpolating new frames between
successive real frames based on object motion.

Motion-based imaging algorithms are traditionally performed in GPUs or CPUs
later in the vision pipeline, but they are increasingly subsumed into ISPs to
improve compute efficiency. Commercial examples of motion-enabled camera ISPs
include ARM Mali C-71~\cite{malicamera}, Qualcomm Spectra ISP~\cite{qcspectraisp}, and products from Hikvision~\cite{hikvision3ddnr}, ASICFPGA~\cite{fpgaasicisp},
PX4FLOW~\cite{px4flow}, Pinnacle Imaging Systems~\cite{denalimcisp}, and
Centeye~\cite{centeye}, just to name a few based on public information.

Today's ISPs generate motion information internally and discard it after the frame
is processed. Instead, we keep the motion information from each frame and expose it at the
SoC-level to inmprove the efficiency of the vision backend.




\subsection{Motion Estimation using Block Matching}
\label{sec:background:mv}

Among various motion estimation algorithms, block-matching (BM)~\cite{bmsurvey}
is widely used in ISP algorithms due to its balance between accuracy and efficiency. Here, we briefly introduce its algorithms and data
structures that we will refer to later.

\begin{figure}[t]
\centering
\subfloat[\small{Block-matching.}]
{
  \includegraphics[trim=0 0 0 0, clip, height=1.6in]{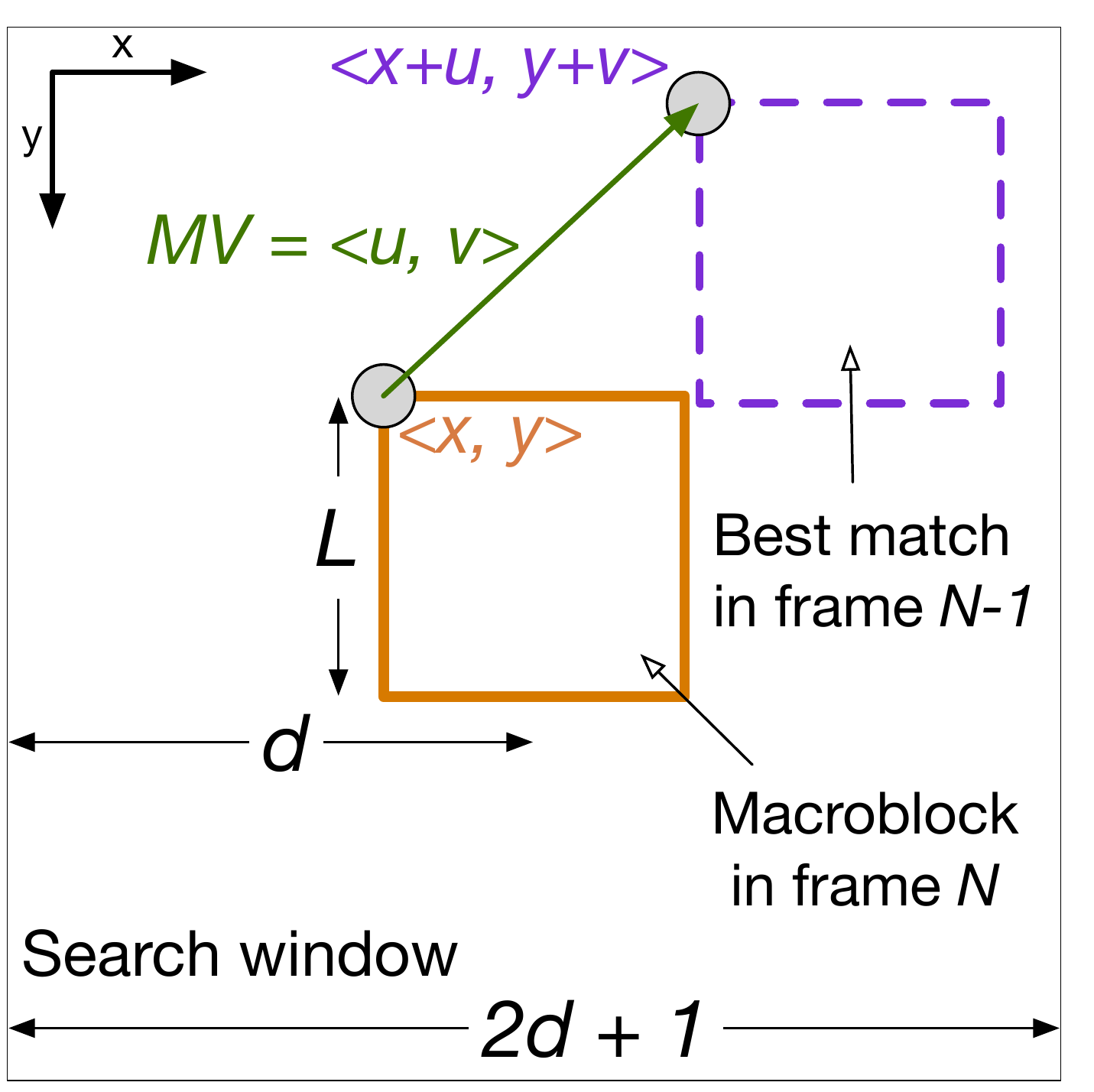}
  \label{fig:bm}
}\hfill
\subfloat[\small{Motion vectors.}]
{
  \includegraphics[trim=0 0 0 0, clip, height=1.57in]{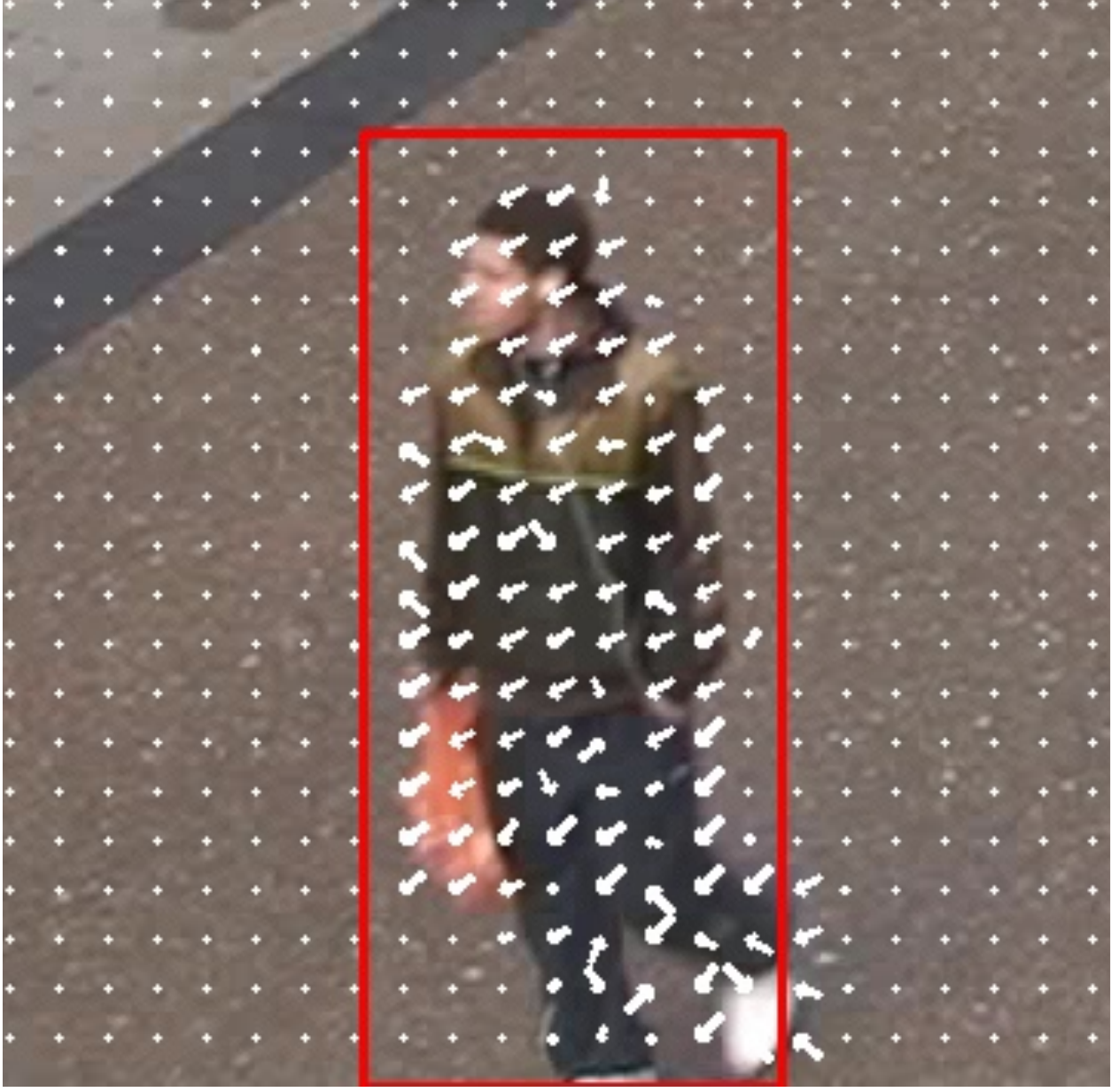}
  \label{fig:mv}
}
\caption{\small Motion estimation. \protect\subref{fig:bm}: Block-matching example in a $(2d+1)\times(2d+1)$ search window. $L$ is the macroblock size. \protect\subref{fig:mv}: Each arrow reprensents an MB's motion vector. MBs in the foreground object have much more prounced motions than the background MBs.}
\label{fig:me}
\end{figure}

The key idea of BM is to divide a frame into multiple $L \times L$ macroblocks (MB), and search in the previous frame for the closest match for each MB using Sum of
Absolute Differences (SAD) of all $L^2$ pixels as the matching metric. The
search is performed within a 2-D search window with $(2d+1)$ pixels in both
vertical and horizontal directions, where $d$ is the search range. \Fig{fig:bm}
illustrates the basic concepts.

Different BM strategies trade-off search accuracy with compute
efficiency. The most accurate approach is to perform an
exhaustive search (ES) within a search window, which requires $L^{2} \cdot
(2d+1)^{2}$ arithmetic operations per MB. Other BM variants trade a small
accuracy loss for computation reduction. For instance, the classic three step search (TSS)~\cite{tss} searches only part of the search window by decreasing $d$ in logarithmic steps. TSS simplifies the amount of
arithmetic operations per MB to $L^{2} \cdot (1+8 \cdot log_{2}(d+1))$, which is
a $8/9$ reduction under $d=7$. We refer interested readers to Jakubowski and Pastuszak~\cite{bmsurvey} for a comprehensive discussion of BM algorithms.

Eventually, BM calculates a \textit{motion vector} (MV) for each MB, which
represents the location offset between the MB and its closest match in the
previous frame as illustrated in~\Fig{fig:bm}. Critically, this offset can be used as an estimation of the MB's 
motion. For instance, an MV <$u, v$> for an MB at location <$x, y$> indicates
that the MB is moved from location <$x+u, y+v$> in the previous frame.
\Fig{fig:mv} visualizes the motion vectors in a frame. Note that
MVs can be encoded efficiently. An MV requires $\lceil
log_{2}(2d+1) \rceil$ bits for each direction, which equates to just 1 byte
of storage under a typical $d$ of seven.

\section{Motion-based Continuous Vision Algorithm}
\label{sec:algo}

The key idea of Euphrates is that pixel changes across frames directly encode object motion. Thus, pixel-level temporal motion information can be
used to simplify continuous vision tasks through motion extrapolation. This
section first provides an overview of the algorithm~(\Sect{sec:algo:ov}). We
then discuss two important aspects of the algorithm: how to
extrapolate~(\Sect{sec:algo:how}) and when to
extrapolate~(\Sect{sec:algo:when}).


\subsection{Overview}
\label{sec:algo:ov}

\begin{figure}[t]
  \centering
  \includegraphics[trim=0 0 0 0, clip, width=.95\columnwidth]{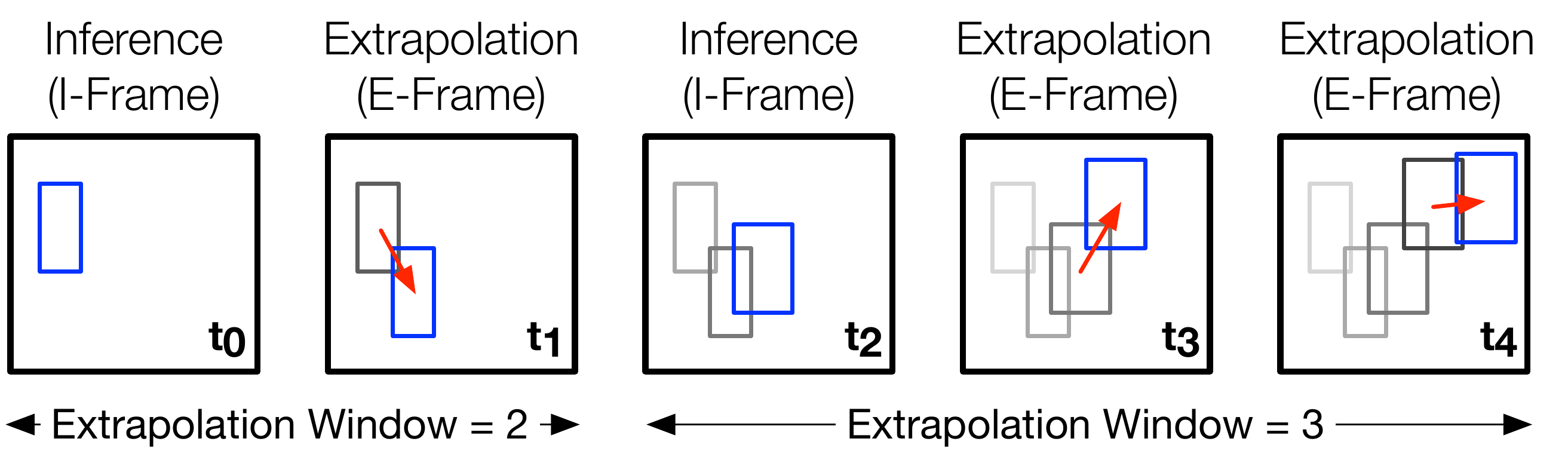}
  \caption{\small{Vision computation results such as Region of Interest (ROIs) are generated using CNN inference in I-frames. ROIs in E-frames are extrapolated from the previous frame using motion information.}}
  \label{fig:algo}
\end{figure}

Euphrates makes a distinction between two frame types: Inference frame (I-frame)
and Extrapolation frame (E-frame). An I-frame refers to a frame where vision
computation such as detection and tracking is executed using expensive CNN
inference with the frame pixel data
as input. 
In contrast, an E-frame refers to a frame where visual results are generated by
extrapolating ROIs from the previous frame, which itself could either be an
I-frame or an E-frame. \Fig{fig:algo} illustrates this process using object
tracking as an example. Rectangles in the figure represent the ROIs of
a single tracked object. Frames at $t_{0}$ and $t_{2}$ are I-frames with ROIs 
generated by full CNN inference. On the other hand, ROIs in frames at $t_{1}$, $t_{3}$, and
$t_{4}$ are extrapolated from their corresponding preceding frames.

Intuitively, increasing the ratio of E-frames to I-frames reduces the
number of costly CNN inferences, thereby enabling higher
frame rates while improving energy-efficiency. 
However, this strategy must have little
accuracy impact to be useful.
As such, the challenge of our algorithm is to strike a balance between
accuracy and efficiency. We identify two aspects that affect the
accuracy-efficiency trade-off: \textit{how} to extrapolate from previous frame,
and \textit{when} to perform extrapolation.


\subsection{How to Extrapolate}
\label{sec:algo:how}

The goal of extrapolation is to estimate the ROI(s) for the current frame without CNN inference by using the motion vectors generated by the ISP. Our hypothesis is that the average motion of all pixels in a visual field can largely estimate the field's global motion. Thus, the first step in the algorithm calculates the average motion vector ($\mu$) for a given ROI according to \Equ{equ:avg_mv}, where $N$ denotes the total number of pixels bounded by the ROI, and $\overrightarrow{v_i}$ denotes the motion vector of the $i^{th}$ bounded pixel. It is important to note that the ISP generates MVs at a macroblock-granularity, and as such each pixel inherits the MV from the MB it belongs to. \Sect{sec:eval:sen} will show that the MV granularity has little impact on accuracy.


\setlength{\abovedisplayskip}{0pt}
\begin{align}
\mu \ &=\ \sum\nolimits_{i}^{N}\overrightarrow{v_i}\ /\ N \label{equ:avg_mv}  \\
\alpha_{F}^{i}\ &=1\ -\ \dfrac{SAD_{F}^{i}}{\ 255 \times L^2} \label{equ:conf} \\
MV_F &= \beta \cdot \mu_{F}\ +\ (1-\beta) \cdot MV_{F-1} \label{equ:mv}
\end{align}

Extrapolating purely based on average motion, however, has two downsides: it is vulnerable to motion vector noise and it does not consider object deformation.

\paragraph{Filtering Noisy Motion Vectors} The block-based motion estimation
algorithm introduces noise in the MVs.
For instance, when a visual object is blurred or occluded (hidden), the block-matching algorithm may
not be able to find a good match within the search window.

To quantify the noise in an MV, we associate a confidence value with each MV.
We empirically find that this confidence is highly correlated with the SAD
value, which is generated during block-matching. Intuitively, a higher SAD value
indicates a lower confidence, and vice versa. \Equ{equ:conf} formulates the
confidence calculation, where $SAD_F^i$ denotes the SAD value of the $i^{th}$
macroblock in frame $F$, and $L$ denotes the dimension of the macroblocks.
Effectively, we normalize an MV's SAD to the maximum possible value (i.e., $255
\times L^2$) and regulate the resultant confidence ($\alpha_{F}^{i}$) to fall
between $[0, 1]$. We then derive the confidence for an ROI by averaging the
confidences of all the MVs encapsulated by the ROI.


The confidence value can be used to filter out the impact of very noisy MVs
associated with a given ROI. This is achieved by assigning a high
weight to the average MV in the current frame ($\mu_{F}$), if its confidence is
high. Otherwise, the motion from previous frames is emphasized. 
In a recursive fashion, this is achieved by directly weighting the contribution
of the current MV against the average~\Equ{equ:mv}, where $MV_F$ denotes the final motion vector
for frame $F$, $MV_{F-1}$ denotes the motion vector for the previous frame, and
$\beta$ is the filter coefficient that is determined by $\alpha$. Empirically,
it is sufficient to use a simple piece-wise function that sets $\beta$ to
$\alpha$ if $\alpha$ is greater than a threshold, and to 0.5 otherwise. The
final motion vector ($MV_F$) is composed with the ROI in the previous
frame to update its new location. That is: $R_F = R_{F-1} + MV_F$.

\begin{figure*}[t]
\vspace{-2pt}
  \begin{minipage}[t]{1.4\columnwidth}
    \centering
    \includegraphics[trim=0 0 0 0, clip, height=1.75in]{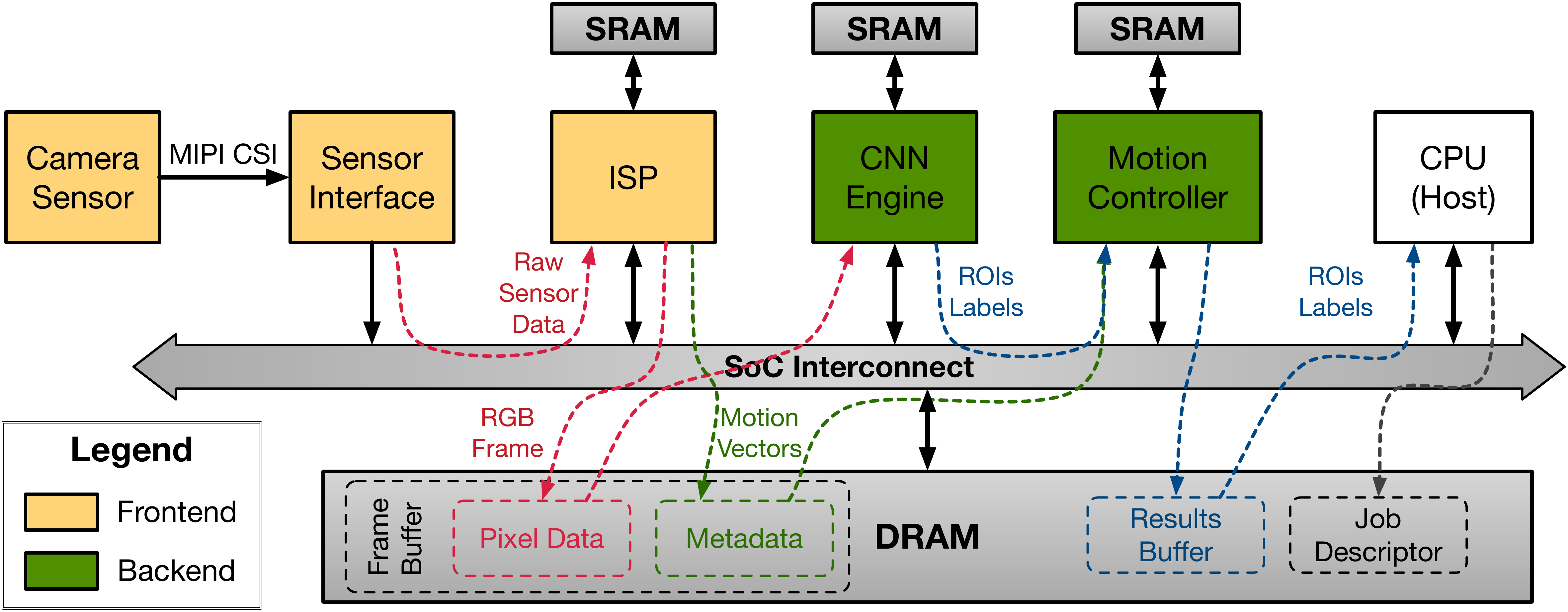}
    \caption{\small{Block diagram of the augmented continuous vision subsystem in a mobile SoC.}}
    \label{fig:soc}
  \end{minipage}
  \hspace{15pt}
  \begin{minipage}[t]{0.5\columnwidth}
    \centering
    \includegraphics[trim=0 0 0 0, clip, height=1.75in]{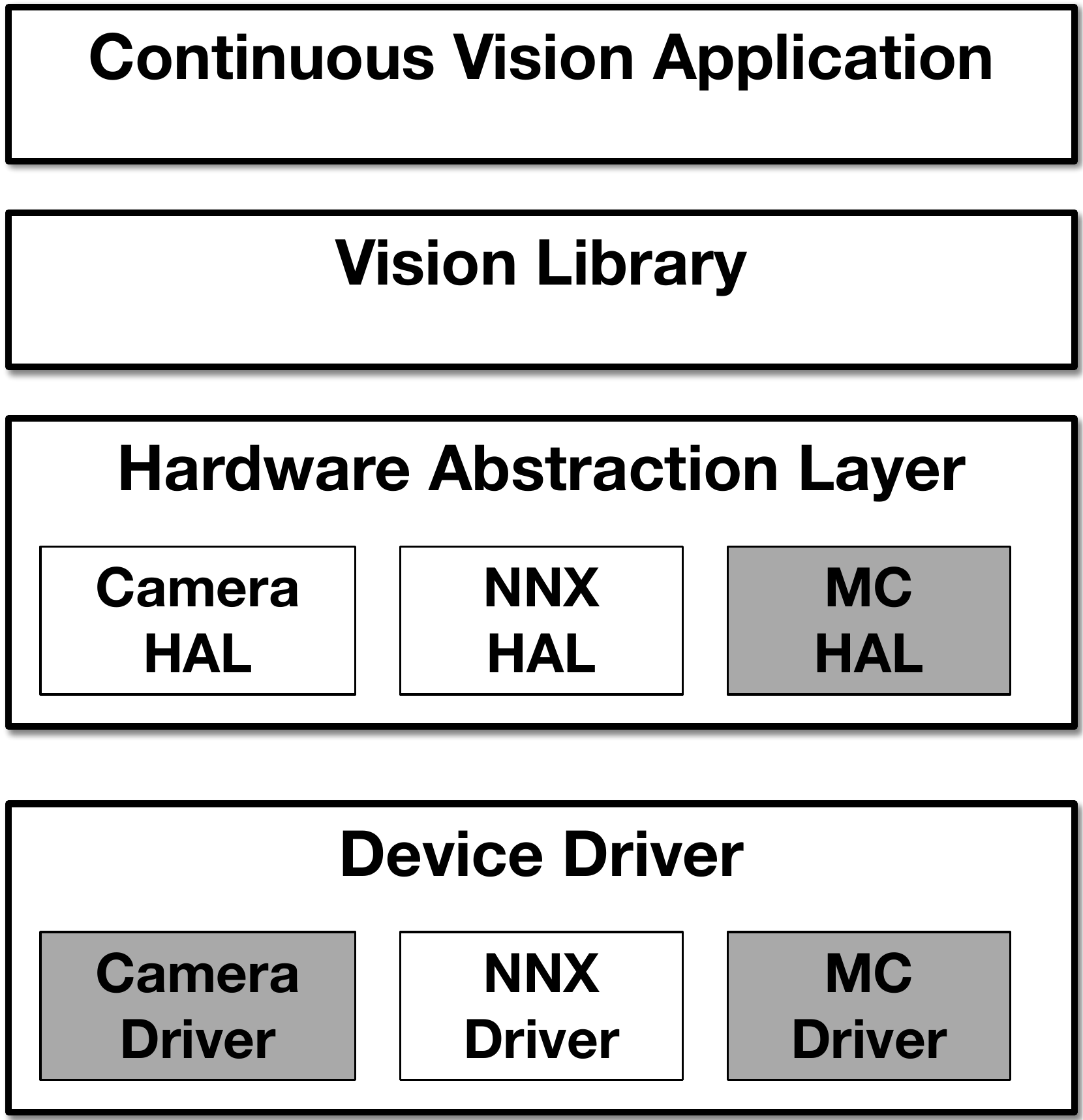}
    \caption{\small{Vision software stack with modifications shaded.}}
    \label{fig:sw}
  \end{minipage}
\end{figure*}

\paragraph{Handle Deformations} Using one global average motion essentially treats an entire scene as a rigid object, which, however, ignores non-rigid deformations. For instance, the head and the arms of a running athlete have different motions that must be taken into account. Inspired by the classic deformable parts model~\cite{dpm}, we divide an ROI into multiple sub-ROIs and apply extrapolation using the above scheme (i.e., \Equ{equ:avg_mv} - \Equ{equ:mv}) for each sub-ROI. In this way, we allow each sub-ROI to move in different directions with different magnitudes. As a result, we get several disconnected sub-ROIs. We then derive the final ROI by calculating the minimal bounding box that encapsulates all the extrapolated sub-ROIs.



\paragraph{Computation Characteristics} Our algorithm is very efficient to
compute. Consider a typical ROI of size 100$\times$50, the
extrapolation step requires only about 10~K 4-bit fixed-point operations per
frame, several orders of magnitude fewer than the billions of operations
required by CNN inferences. 



\subsection{When to Extrapolate}
\label{sec:algo:when}

Another important aspect of the extrapolation algorithm is to decide which
frames to execute CNN inference on, and which frames to extrapolate. To simplify
the discussion, we introduce the notion of an Extrapolation Window (EW), which is the number
of consecutive frames between two I-frames (exclusive) as shown in~\Fig{fig:algo}. Intuitively, as
EW increases, the compute efficiency improves, but errors introduced by
extrapolation also start accumulating, and vice versa. Therefore, EW is an
important knob that determines the trade-off between compute efficiency and
accuracy. Euphrates provides two modes for setting EW: constant mode and
adaptive mode.

Constant mode sets EW statically to a fixed value. This provides predictable,
bounded performance and energy-efficiency improvements. 
For instance, under $EW=2$, one can 
estimate that the amount of computation per frame is reduced by half,
translating to 2$\times$ performance increase or 50\% energy savings.


However, the constant mode can not adapt to extrapolation inaccuracies. For
instance, when an object partially enters the frame, the block-matching
algorithm will either not be able to find a good match within its search window
or find a numerically-matched MB, which, however, does not represent the
actual motion. In such case, CNN inference can provide a more accurate vision
computation result.

We introduce a dynamic control mechanism to respond
to inaccuracies introduced by motion extrapolation. Specifically, whenever a CNN
inference is triggered, we compare its results with those obtained from
extrapolation. If the difference is larger than a threshold, we
incrementally reduce EW; similarly, if the difference is consistently lower
than the threshold across several CNN invocations, we incrementally increase
EW.

The two modes are effective in different scenarios. The constant mode is useful when
facing a hard energy or frame rate bound. When free of such constraints,
adaptive mode improves compute efficiency with little accuracy
loss.



\section{Architecture Support}
\label{sec:arch}

In this section, we start from a state-of-the-art mobile SoC, and show how to
co-design the SoC architecture with the proposed algorithm. After explaining our
design philosophy and providing an overview~(\Sect{sec:arch:ov}), we describe
the hardware augmentations required in the frontend~(\Sect{sec:arch:fe}) and
backend of the vision subsystem~(\Sect{sec:arch:be}). Finally, we discuss the
software implications of our architecture extensions~(\Sect{sec:arch:sw}).

\subsection{Design Philosophy and System Overview}
\label{sec:arch:ov}

\paragraph{Design Principles} Two principles guide our SoC design. First, the vision pipeline in the SoC
must act autonomously, to avoid constantly interrupting the CPU which needlessly
burns CPU cycles and power~(\Sect{sec:background:pipe}). This design
principle motivates us to provide SoC architecture support, rather than
implementing the new algorithm in software, because the latter would involve the CPU
in every frame.


Second, the architectural support for the extrapolation functionality should be
decoupled from CNN inference. This design principle motivates
us to propose a separate IP to support the new functionality rather than
augmenting an existing CNN accelerator. The rationale is that CNN accelerators
are still evolving rapidly with new models and architectures constantly
emerging. Tightly coupling our algorithm with any particular CNN accelerator is
inflexible in the long term. Our partitioning accommodates future changes in
inference algorithm, hardware IP (accelerator, GPU, etc.), or even IP vendor.


\paragraph{System Overview} \Fig{fig:soc} illustrates the augmented mobile SoC architecture. In
particular, we propose two architectural extensions. First, motivated by the
synergy between the various motion-enabled imaging algorithms in the ISP and our
motion extrapolation CV algorithm, we augment the ISP to expose the motion
vectors to the vision backend. Second, to coordinate the backend under the new
algorithm without significant CPU intervention, we propose a new hardware IP called
the motion controller. The frontend and backend communicate through the system
interconnect and DRAM.




Our proposed system works in the following way. The CPU initially configures the
IPs in the vision pipeline, and initiates a vision task by writing a job
descriptor.
The camera sensor module captures real-time raw images, which are fed
into the ISP. The ISP generates, for each frame, both pixel data and metadata
that are transferred to an allocated frame buffer in DRAM. The motion vectors and the
corresponding confidence data are packed as part of the metadata in the frame
buffer.

The motion controller sequences operations in the backend and coordinates with
the CNN engine. It directs the CNN engine to read image pixel data to execute an
inference pass for each I-frame. The inference results, such as predicted ROIs
and possibly classification labels for detected objects, are written to
dedicated memory mapped registers in the motion controller through the system
interconnect. The motion controller combines the CNN inference data and the
motion vector data to extrapolate the results for E-frames.


\begin{figure}[t]
  \centering
  \includegraphics[trim=0 0 0 0, clip, width=\columnwidth]{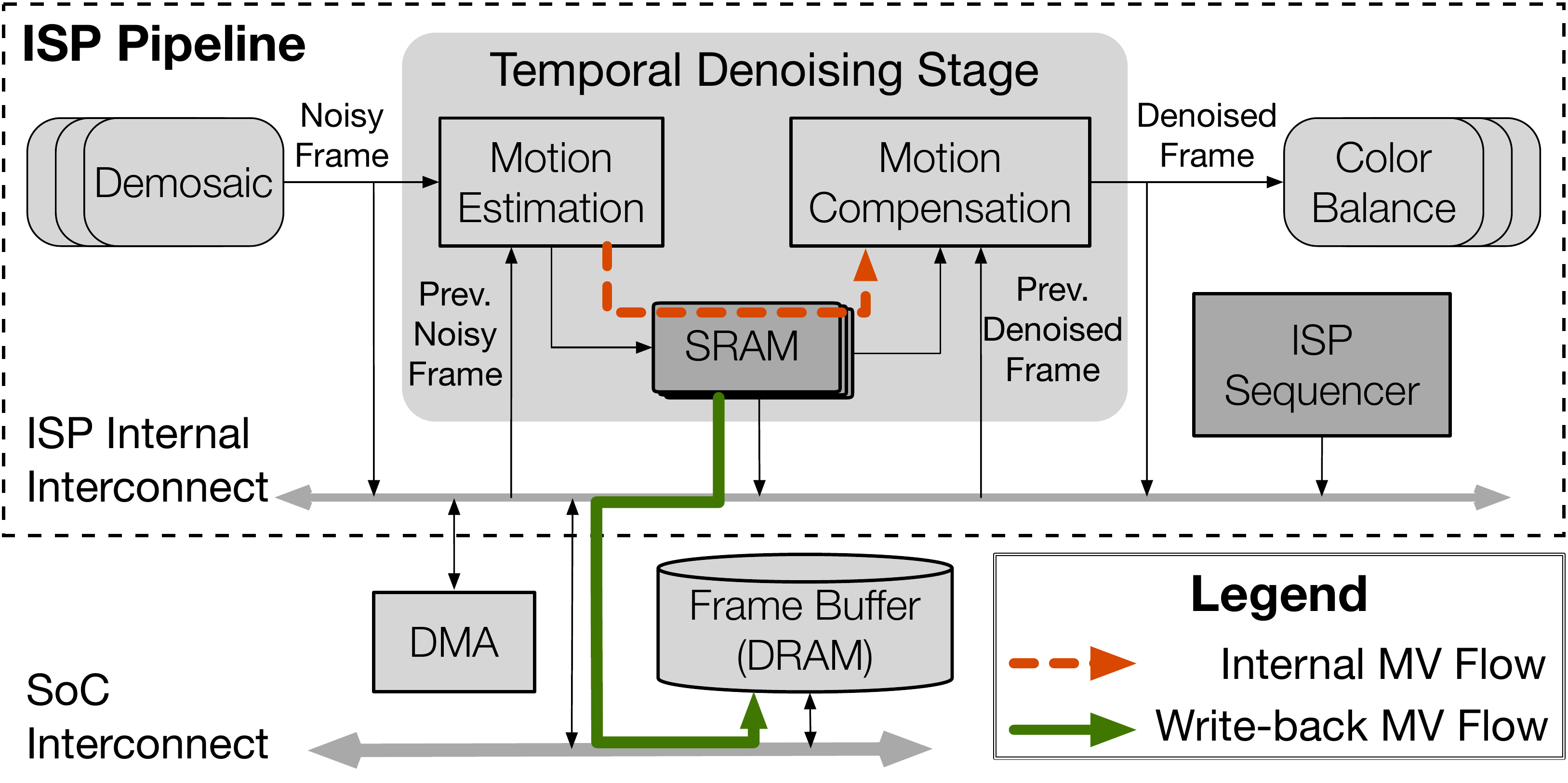}
  \caption{\small{Euphrates augments the ISP to expose the motion vectors to the rest of the SoC with lightweight hardware extensions. The motion vector traffic is off the critical path and has little performance impact.}}
  \label{fig:isp}
\end{figure}

\subsection{Augmenting the Vision Frontend}
\label{sec:arch:fe}

Motion vectors (MVs) are usually reused internally within an ISP and are
discarded after relevant frames are produced. Euphrates exposes MVs to the
vision backend, which reduces the overall computation and simplifies the
hardware design.

To simplify the discussion, we assume that the motion vectors are generated
by the temporal denoising (TD) stage in an ISP. \Fig{fig:isp} illustrates how
the existing ISP pipeline is augmented. The motion estimation block in the TD
stage calculates the MVs and buffers them in a small local SRAM.
The motion compensation block then uses the MVs to denoise the current frame. After
the current frame is temporally-denoised, the corresponding SRAM space can be
recycled. In augmenting the ISP pipeline to expose the MV data, we must decide:
1) by what means the MVs are exposed to the system with minimal design cost, and
2) how to minimize the performance impact on the ISP pipeline.


\paragraph{Piggybacking the Frame Buffer} We propose to expose the
MVs by storing them in the metadata section of the frame buffer, which resides
in the DRAM and is accessed by other SoC IPs through the existing system memory
management unit. This augmentation is implemented by modifying the ISP's
sequencer to properly configure the DMA engine.

Piggybacking the existing frame buffer mechanism rather than adding a dedicated
link between the ISP and the vision backend has the minimum design cost with
negligible memory traffic overhead. Specifically, a 1080p frame (1920 $\times$
1080) with a 16 $\times$ 16 macroblock size will produce 8,100 motion vectors,
equivalent to only about 8~KB per frame (Recall from~\Sect{sec:background:mv} that
each motion vector can be encoded in one byte), which is a very small fraction
of the 6~MB frame pixel data that is already committed to the frame buffer.


\paragraph{Taking MV Traffic off the Critical-path}  A naive design to handle MV
write-back might reuse the existing local SRAM in the TD stage as the DMA
buffer. However, this  strategy would stall the ISP pipeline due to SRAM
resource contention. This is because the ISP's local SRAMs are typically 
sized to precisely for the data storage required, thanks to the deterministic
data-flow in imaging algorithms. Instead, to take the MV write traffic off the critical
path, we propose to double-buffer the SRAM in the TD stage at a slight cost in area
overhead. In this way, the DMA engine opportunistically initiates the MV
write-back traffic as it sees fit, effectively overlapping the write-back with
the rest of the ISP pipeline.




\subsection{Augmenting the Vision Backend}
\label{sec:arch:be}

\begin{figure}[t]
  \centering
  \includegraphics[trim=0 0 0 0, clip, width=\columnwidth]{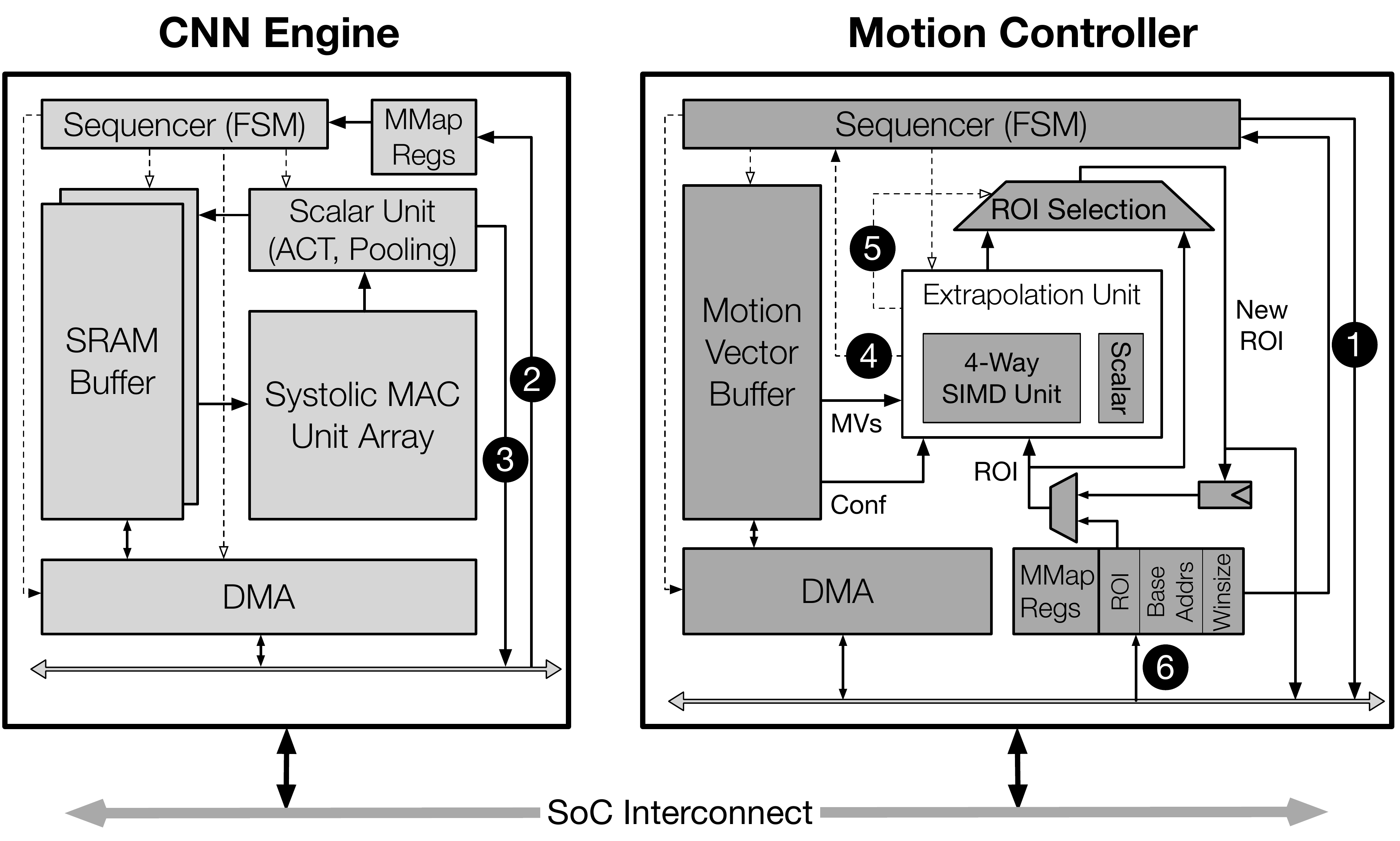}
  \caption{\small{Euphrates adds the motion controller to the vision backend, alongside an existing, unmodified CNN inference accelerator. Dash lines are control signals and solid lines represent the data flow.}}
  \label{fig:arch}
\end{figure}

We augment the vision backend with a new IP called the motion controller. It's job is two-fold. First, it executes the
motion extrapolation algorithm. Second, it coordinates the CNN engine without interrupting the CPU. 
The CNN accelerator is left unmodified, with the same interface to the SoC interconnect.

We design the motion controller engine as a micro-controller ($\mu C$) based IP,
similar to many sensor co-processors such as Apple's Motion
Co-processor~\cite{applemotioncoproc}. It sits on the system interconnect,
alongside the CNN accelerator. The main difference between our IP and a
conventional $\mu C$ such as ARM Cortex-M series~\cite{cortexm} is that the
latter typically does not have on-chip caches. Instead, our extrapolation engine
is equipped with an on-chip SRAM that stores the motion vectors and is fed by a
DMA engine. In addition, the IP replaces the conventional
instruction fetch and decode mechanisms with a programmable sequencer, which
reduces energy and area, while still providing programmability to control the
datapath.

\Fig{fig:arch} shows the microarchitecture of the extrapolation controller.
Important data and control flows in the figure are numbered. The
motion controller is assigned the master role and the CNN engine acts as
a slave in the system. The
master IP controls the slave IP by using its sequencer to program the slave's
memory-mapped registers (\circled{white}{1} and \circled{white}{2} in
\Fig{fig:arch}). The slave IP always returns the computation results to the
master IP (\circled{white}{3}) instead of directly interacting with the CPU. We
choose this master-slave separation, instead of the other way around, because it
allows us to implement all the control logics such as adaptive EW completely in
the extrapolator engine without making assumptions about the CNN accelerator's
internals.


The core of the motion controller's datapath is an extrapolation unit which includes a SIMD unit and a scalar unit. The extrapolation operation is highly parallel~(\Sect{sec:algo:how}), making SIMD a nature fit. The scalar unit is primarily responsible for generating two signals: one that controls the EW size in the adaptive mode (\circled{white}{4}) and the other that chooses between inferenced and extrapolated results (\circled{white}{5}). The IP also has a set of memory-mapped registers that are programmed by the CPU initially and receive CNN engine's inference results (\circled{white}{6}).

\subsection{Software Implications}
\label{sec:arch:sw}

Euphrates makes no changes to application developers' programming interface and
the CV libraries. This enables software backward
compatibility. Modifications to the Hardware Abstraction Layer (HAL) and device
drivers are required.

We show the computer vision software stack in \Fig{fig:sw} with enhancements shaded. The new motion controller (MC) needs to be supported at both the HAL and driver layer. In addition, the camera driver is enhanced to configure the base address of motion vectors. However, the camera HAL is left intact because motion vectors need not be visible to OS and programmers. We note that the CNN engine (NNX) driver and HAL are also unmodified because of our design decision to assign the master role to the motion controller IP.



\section{Implementation and Experimental Setup}
\label{sec:exp}

This section introduces our hardware modeling methodology (\Sect{sec:exp:hw})
and software infrastructure (\Sect{sec:exp:workloads}). 

\subsection{Hardware Setup}
\label{sec:exp:hw}


We develop an in-house simulator with a methodology similar to the GemDroid~\cite{gemdroid} SoC simulator. The simulator includes a functional model, a performance model, and an power model for evaluating the continuous vision pipeline. The functional model
takes in video streams to mimic real-time camera capture and implements the
extrapolation algorithm in OpenCV, from which we derive accuracy results. The
performance model captures the timing behaviors of various vision pipeline components including the camera
sensor, the ISP, the CNN accelerator, and the motion controller. It then models the timing of cross-IP activities, from
which we tabulate SoC events that are fed into the power model for energy
estimation.

Whenever possible, we calibrate the power model by measuring the Nvidia
Jetson TX2 module~\cite{tx2}, which is widely used in mobile vision
systems. TX2 exposes several SoC and board level power rails to a Texas Instruments INA 3221 voltage monitor IC, from which we retrieve power consumptions through the I2C interface. Specifically, the ISP power is obtained from taking the differential power at the \textsf{VDD\_SYS\_SOC} power rail between idle and active mode, and the main memory power is obtained through the \textsf{VDD\_SYS\_DDR} power rail.
We develop RTL models
or refer to public data sheets when direct measurement is unavailable. Below we
discuss how major components are modeled.





\begin{table}[t]
\vspace{-3pt}
\centering
\Huge
\caption{\small Details about the modeled vision SoC.}
\label{my-label}
\renewcommand*{\arraystretch}{1.1}
\renewcommand*{\tabcolsep}{13pt}
\resizebox{\columnwidth}{!}
{
  \begin{tabular}{ll}
  \toprule[0.15em]
  \textbf{Component}         & \textbf{Specification}\\
  \midrule[0.05em]
  Camera Sensor              & ON Semi AR1335, 1080p @ 60 FPS\\
  \midrule[0.05em]
  ISP                        & 768~MHz, 1080p @ 60 FPS\\
  \midrule[0.05em]
  \specialcell{NN Accelerator\\(NNX)}             & \specialcell{
                              $24\times24$ systolic MAC array\\ 
                              1.5 MB double-buffered local SRAM\\
                              3-channel, 128bit AXI4 DMA Engine} \\
  \midrule[0.05em]
  \specialcell{Motion Controller\\(MC)}        & \specialcell{
                              4-wide SIMD datapath\\
                              8KB local SRAM buffer\\
                              3-channel, 128bit AXI4 DMA Engine} \\

  \midrule[0.05em]
  DRAM                       & 4-channel LPDDR3, 25.6~GB/s peak BW\\
  \bottomrule[0.15em]
  \end{tabular}
}
\end{table}

\paragraph{Image Sensor} We model the AR1335~\cite{ar1335}, an image sensor used in many mobile vision systems including the Nvidia Jetson TX1/TX2 modules~\cite{eCAM131}. In our evaluation, we primarily consider the common sensing setting that captures 1920
$\times$ 1080 (1080p) videos at 60 FPS. AR1335 is estimated to consume 180~mW of power under this setting~\cite{ar1335}, representative of common digital camera sensors available on the
market~\cite{ov5693, ar0542, mt9p031}. We directly integrate the power consumptions reported in the data sheet into our modeling infrastructure.

\paragraph{ISP} We base our ISP modeling on the specifications of the ISP
carried on the Jetson TX2 board. We do not model the ISP's detailed
microarchitecture but capture enough information about its memory traffic. We
make this modeling decision because our lightweight ISP modification has little
effect on the ISP datapath, but does impact memory
traffic~(\Sect{sec:arch:fe}). The simulator generates ISP memory traces given a particular resolution and frame rate. The
memory
traces are then interfaced with the DRAM simulator as described
later.


We take direct measurement of the ISP on TX2 for power estimation. Under 1080p
resolution at 60 FPS, the ISP is measured to consume 153~mW, comparable to other
industrial ISPs~\cite{ap0101cs}. We could not obtain any public
information as to whether the TX2 ISP 
performs motion estimation. According
to~\Sect{sec:background:mv}, a 1080p image requires about 50 million arithmetic
operations to generate motion vectors, which is about 2.5\% compute overhead
compared to a research ISP~\cite{darkroom}, and will be much smaller compared
to commercial ISPs. We thus conservatively
factor in 2.5\% additional power.




\paragraph{Neural Network Accelerator} We develop a systolic array-based CNN accelerator
and integrate it into our evaluation infrastructure. The design is reminiscent
of the Google Tensor Processing Unit (TPU)~\cite{tpu}, but is much smaller, as
befits the mobile budget~\cite{dlforarchitects}.

The accelerator consists of a $24\times24$ fully-pipelined
Multiply-Accumulate array clocked at 1GHz, representing a raw peak
throughput of 1.152 TOPS. A unified, double-buffered SRAM array holds both weights and
activations and is 1.5 MB in total.  A scalar unit is used to handle
per-activation tasks such as scaling, normalization, activation, and pooling.


We implement the accelerator in RTL and synthesize, place, and route the design using 
Synposys and Cadence tools in a 16nm process
technology. Post-layout results show an area of 1.58 $mm^{2}$ and a power consumption of 651~mW. This is equivalent to a power-efficiency of
1.77~TOPS/W, commensurate with recent
mobile-class CNN accelerators that demonstrate around 1\textasciitilde 3
TOPs/W~\cite{myriad2, kaistioe, kaistulp, stmdcnn, envision, eyerisschip} in silicon.

To facilitate future research, we open-source our cycle-accurate simulator of the systolic array-based DNN accelerator, which can provide
performance, power and area requirements for a parameterized accelerator on a
given CNN model~\cite{scalesim}.



\paragraph{Motion Controller} 
The motion controller has a light compute requirement, and is implemented as a 4-wide SIMD datapath with 8~KB local
data SRAM. The SRAM is sized to hold the motion vectors for one 1080p frame with a $16\times 16$ MB size. The IP is clocked at 100~MHz to support 
10 ROIs per frame at 60 FPS, sufficient to cover the peak case in our datasets.
We implement the motion controller IP in RTL, and use the same tool chain and
16nm process technology.
The power consumption is 2.2~mW, which is just slightly
more than a typical micro-controller that has SIMD support (e.g., ARM M4~\cite{armm4trm}). The area is negligible (35,000
$um^{2}$).

\paragraph{DRAM} We use DRAMPower~\cite{drampower} for power estimation. We
model the specification of an 8GB memory with 128-bit interface,
similar to the one used on the Nvidia Jetson TX2. We further validate the simulated
DRAM power consumption against the hardware measurement obtained
from the Jetson TX2 board. Under
the 1080p and 60 FPS camera capture setting, the DRAM consumes about 230~mW.


\begin{table}[t]
\vspace{-3pt}
\Huge
\centering
\caption{\small Summary of benchmarks. GOPS for each neural network is estimated under the 60 FPS requirement. Note that our baseline CNN accelerator provides 1.15 TOPS peak compute capability.}
\renewcommand*{\arraystretch}{1.1}
\renewcommand*{\tabcolsep}{12pt}
\resizebox{\columnwidth}{!}
{
  \begin{tabular}{lllll}
  \toprule[0.15em]
  \textbf{\specialcell{Application \\Domain}} & \textbf{\specialcell{Neural\\Network}} & \textbf{GOPS} & \textbf{Benchmark} & \textbf{\specialcell{Total \\Frames}}\\
  \midrule[0.05em]
  \multirow{2}{*}{\specialcell{Object \\Detection}} & Tiny YOLO & ~~~675  & \multirow{2}{*}{\specialcell{In-house Video \\Sequences}} & \multirow{2}{*}{~~7,264}\\
  ~                                 & YOLOv2   & ~3423 & ~                                         & ~\\
  \midrule[0.05em]
  \multirow{2}{*}{\specialcell{Object \\Tracking}} & \multirow{2}{*}{MDNet} & \multirow{2}{*}{~~~635} & OTB 100 & 59,040\\
  ~                  & ~               & ~                                & VOT 2014               &  10,213 \\
  \bottomrule[0.15em]
  \end{tabular}
}
\label{tab:workloads}
\end{table}

\subsection{Software Setup}
\label{sec:exp:workloads}

\begin{figure*}[t]
\centering
\subfloat[\small{Average precision comparison.}]
{
  \includegraphics[trim=0 0 0 0, clip, height=1.7in]{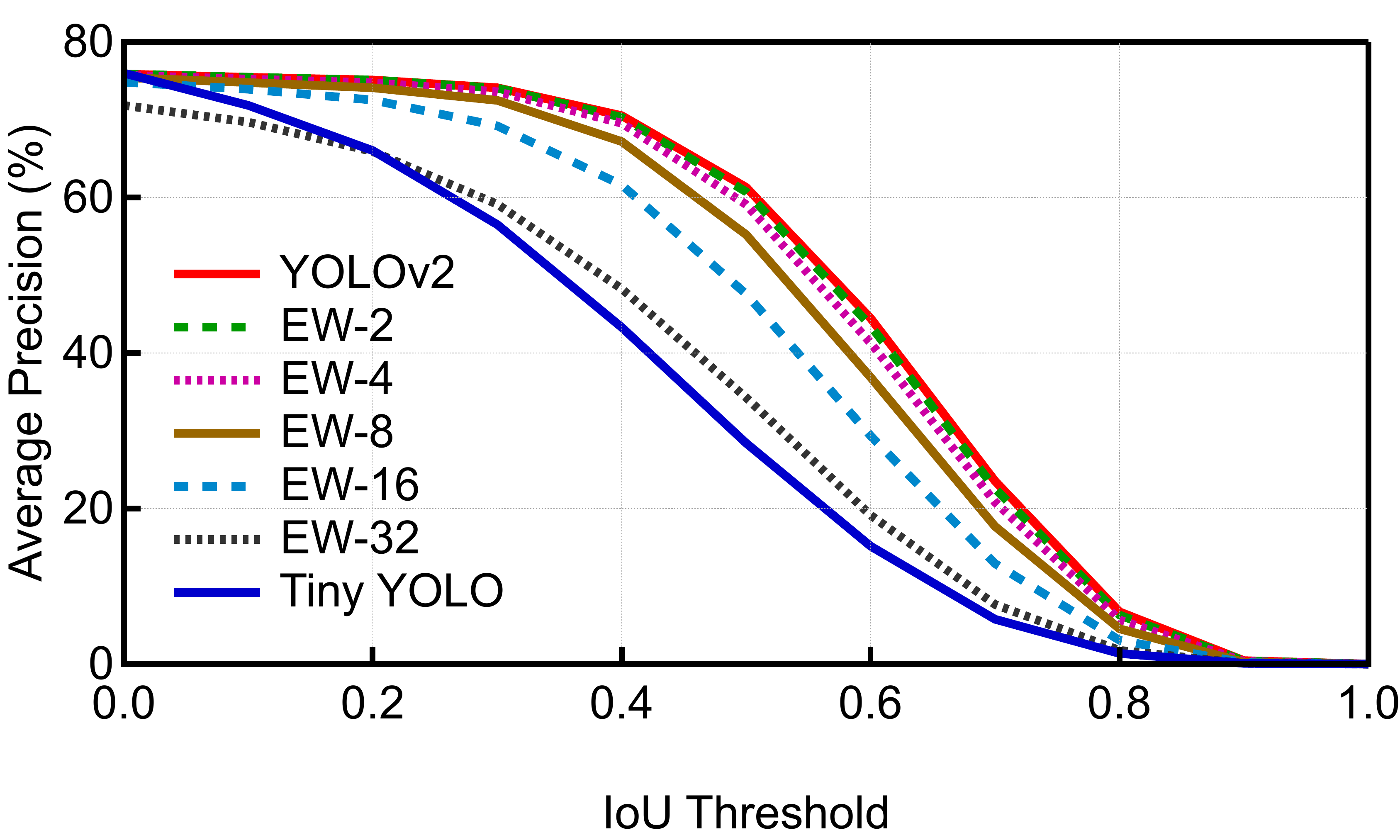}
  \label{fig:yolo_ped_accuracy}
}\hfill
\subfloat[\small{Energy and FPS comparison.}]
{
  \includegraphics[trim=0 0 0 0, clip, height=1.7in]{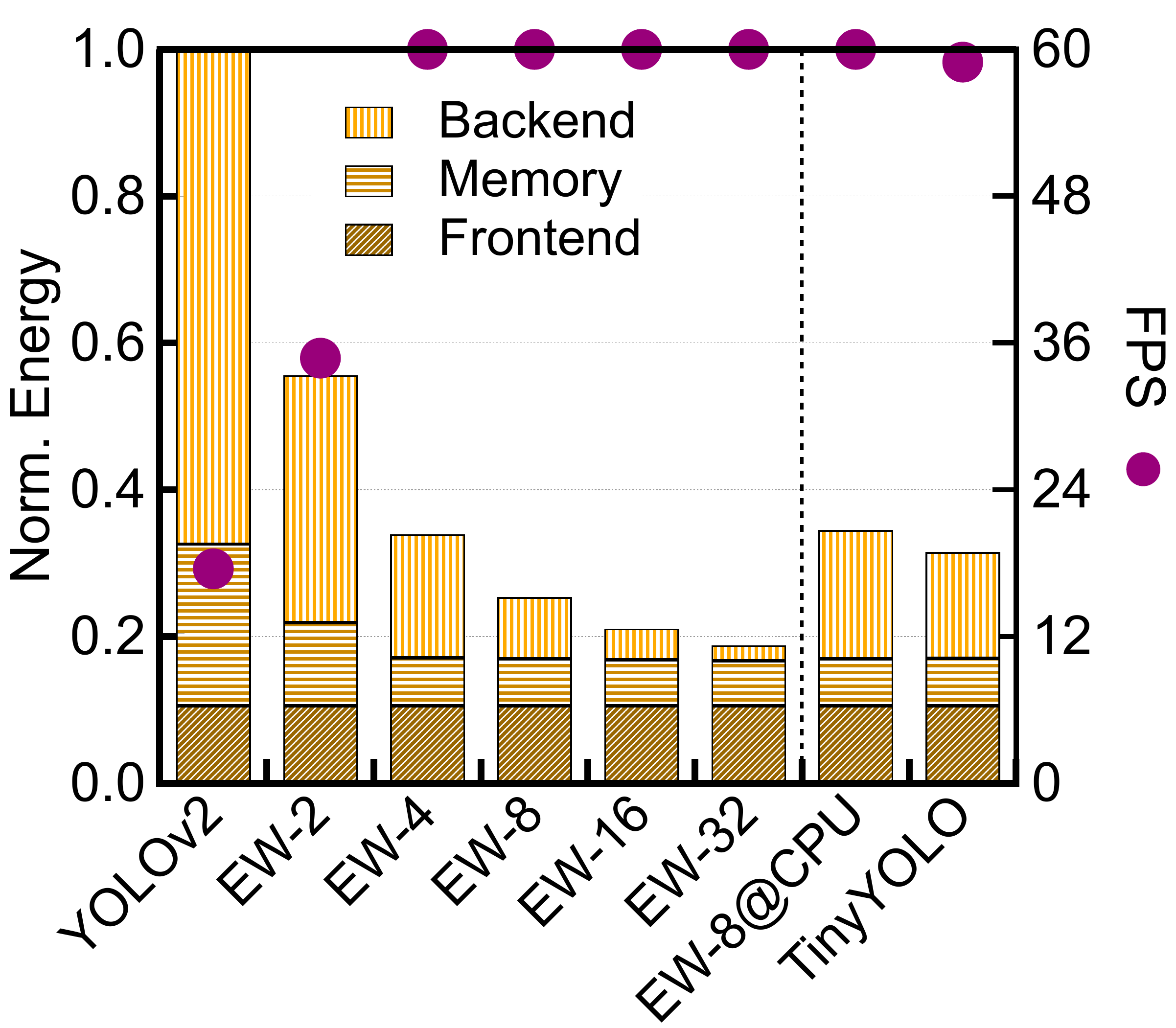}
  \label{fig:yolo_ped_energy}
}\hfill
\subfloat[\small{Compute and memory comparison.}]
{
  \includegraphics[trim=0 0 0 0, clip, height=1.7in]{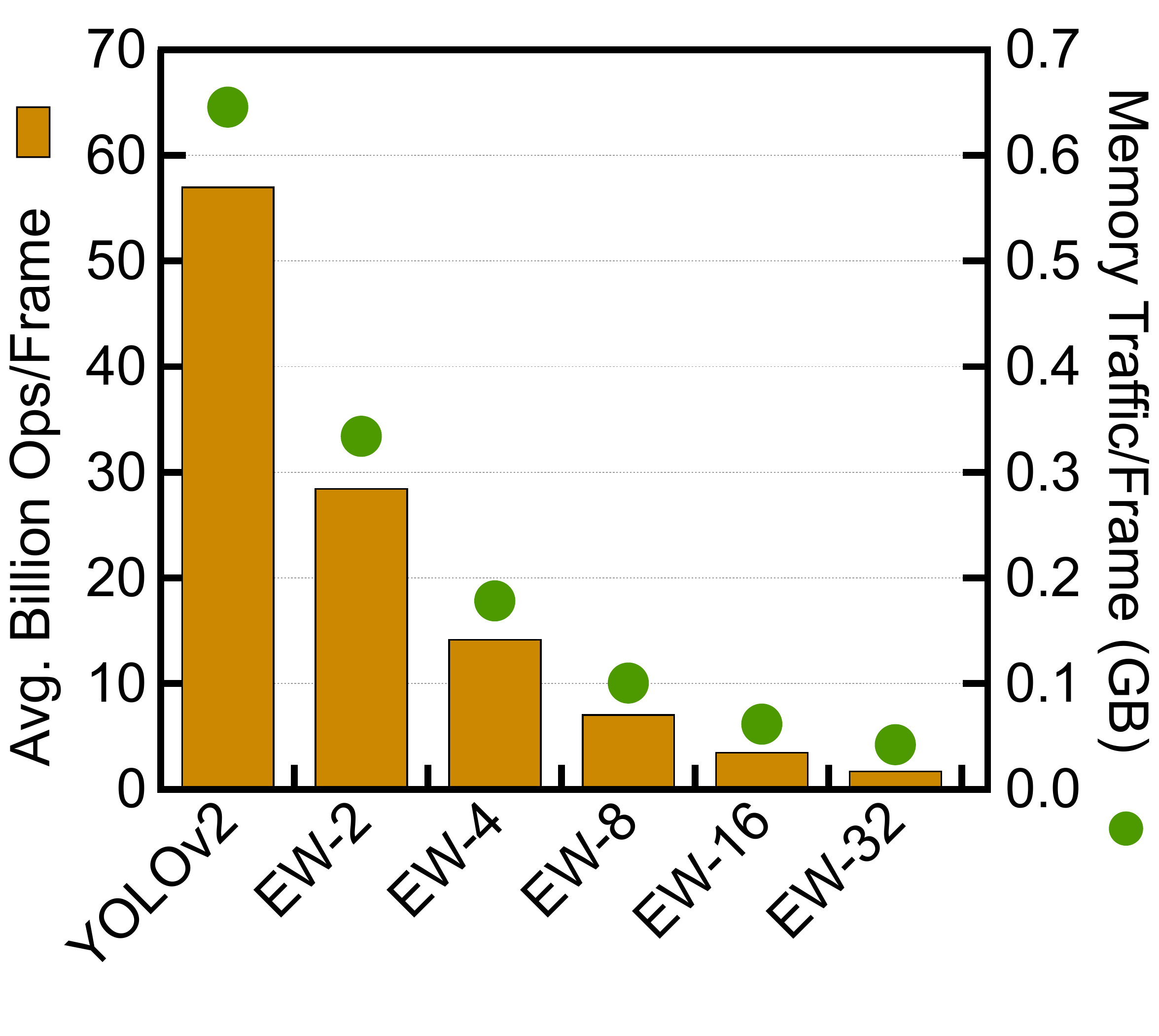}
  \label{fig:yolo_ped_char}
}
\caption{\small Average precision, normalized energy consumption, and FPS
comparisons between various object detection schemes. Energy is broken-down into three main components: backend (CNN engine and motion controller), main memory, and frontend (sensor and ISP).}
\label{fig:yolo_ped}
\end{figure*}


We evaluate Euphrates under two popular mobile continuous vision scenarios:
object detection and object tracking. Their corresponding workload setup is summarized
in~\Tbl{tab:workloads}.

\paragraph{Object Detection Scenario} 
In our evaluation, we study a state-of-the-art object detection CNN called
YOLOv2~\cite{yolo, yolov2}, which achieves the best accuracy and performance
among all the object detectors. As \Tbl{tab:workloads} shows, YOLOv2 requires
over 3.4 TOPS compute capability at 60 FPS,
significantly exceeding the mobile compute budget. For comparison
purposes, we also evaluate a scaled-down version of YOLOv2 called Tiny YOLO. At the
cost of 20\% accuracy loss~\cite{yolowebsite}, Tiny YOLO reduces the
compute requirement to 675 GOPS, which is within the capability of our CNN accelerator.





We evaluate object detection using an in-house video dataset.
We could not use public object detection benchmarks (such as Pascal VOC
2007~\cite{voc2007}) because they are mostly composed of standalone images
without temporal correlation as in real-time video streams. Instead, we capture
a series of videos and extract image sequences. Each image is then manually
annotated with bounding boxes and labels. The types of object classes are
similar to the ones in Pascal VOC 2007 dataset. Overall, each frame contains
about 6 objects and the whole dataset includes 7,264 frames, similar to the
scale of Pascal VOC 2007 dataset. We plan to release the dataset in the future.

We use the standard Intersect-over-Union (IoU) score as an accuracy
metric for object detection~\cite{voc, mscoco}. IoU is the ratio between the intersection
and the union area between the predicted ROI and the ground truth. A detection
is regarded as a true positive (TP) if the IoU
value is above a certain threshold; otherwise it is regarded as a false positive
(FP). The final detection accuracy is evaluated as $TP/(TP + FP)$ across
all detections in all frames, also known as the average precision (AP) score in
object detection literature~\cite{voc}.



\paragraph{Visual Tracking Scenario} We evaluate a state-of-the-art, CNN-based tracker called
MDNet~\cite{mdnet}, which is the winner of the Video Object Tracking (VOT)
challenge~\cite{vot2014benchmark}.
\Tbl{tab:workloads} shows that compared to object detection, tracking is less
compute-intensive and can achieve 60 FPS using our CNN accelerator. However, many
visual tracking scenarios such
as autonomous drones and video surveillance do not have active cooling.
Thus, there is an increasing need to reduce the power/energy consumption of
visual tracking~\cite{adaschallenges}.


We evaluate two widely used object tracking benchmarks:
Object Tracking Benchmark (OTB) 100~\cite{otb100, otbdataset} and VOT
2014~\cite{vot2014, vot2014benchmark}. OTB 100 contains 100 
videos with different visual attributes such as illumination variation and occlusion that mimic realistic tracking scenarios in the wild. VOT
2014 contains 25 sequences with irregular bounding boxes, complementing the OTB
100 dataset. In total, we evaluate about 70,000 frames. We use the standard
success rate as the accuracy metric~\cite{otb}, which represents the percentage
of detections that have an IoU ratio above a certain threshold.

\section{Evaluation}
\label{sec:eval}

We first quantify the effectiveness of Euphrates under object
detection~(\Sect{sec:eval:detect}) and 
tracking~(\Sect{sec:eval:tracking}) scenarios. We then show that Euphrates is robust against motion estimation results
produced in the vision frontend~(\Sect{sec:eval:sen}).

%

\subsection{Object Detection Results}
\label{sec:eval:detect}

Euphrates doubles the achieved FPS with 45\% energy saving at the
cost of only 0.58\% accuracy loss. Compared to the conventional
approach of reducing the CNN compute cost by scaling down the network
size, Euphrates achieves a higher frame rate,
lower energy consumption, and higher accuracy.

\paragraph{Accuracy Results} \Fig{fig:yolo_ped_accuracy} compares the average
precision (AP) between baseline YOLOv2 and Euphrates under different
extrapolation window sizes (EW-N, where N ranges from 2 to 32 in powers
of 2). For a comprehensive comparison, we vary the IoU ratio from 0 (no overlap)
to 1 (perfect overlap). Each <$x, y$> point corresponds to the percentage of
detections ($y$) that are above a given IoU ratio ($x$). Overall, the AP
declines as the IoU ratio increases.

Replacing expensive NN inference with cheap motion extrapolation has
negligible accuracy loss. EW-2 and EW-4 both achieve a success rate close to the
baseline YOLOv2, represented by the close proximity of their corresponding curves
in \Fig{fig:yolo_ped_accuracy}. Specifically, under an IoU of 0.5, which is
commonly regarded as an acceptable detection threshold~\cite{vocretro}, EW-2
loses only 0.58\% accuracy compared to the baseline YOLOv2.

\paragraph{Energy and Performance} The energy savings and FPS improvements are
significant. \Fig{fig:yolo_ped_energy} shows the energy consumptions of
different mechanisms normalized to the baseline YOLOv2. We also overlay the FPS
results on the right $y$-axis. The energy consumption is split into three
parts: frontend (sensor and ISP), main memory, and backend (CNN engine and
motion controller). The vision frontend is configured to produce frames at a constant
60 FPS in this experiment. Thus, the frontend energy is the same across
different schemes.

The baseline YOLOv2 consumes the highest energy and can only achieve about 17
FPS, which is far from real-time. As we increase EW, the total energy
consumption drops and the FPS improves. Specifically, EW-2 reduces the total
energy consumption by 45\% and improves the frame rate from 17 to 35; EW-4
reduces the energy by 66\% and achieves real-time frame rate at 60 FPS.
The frame rate caps at EW-4, limited by the frontend. Extrapolating beyond
eight consecutive frames have higher accuracy loss with only marginal energy
improvements. This is because as EW size increases the energy consumption
becomes dominated by the vision frontend and memory.

The significant energy efficiency and performance improvements come from two
sources: relaxing the compute in the backend and reducing the SoC memory
traffic. \Fig{fig:yolo_ped_char} shows the amount of arithmetic operations and
SoC-level memory traffic (both reads and writes) per frame under various
Euphrates settings. As EW increases, more expensive CNN inferences are replaced
with cheap extrapolations (\Sect{sec:algo:how}), resulting in
significant energy savings. Euphrates also reduces the amount of SoC memory
traffic. This is because E-frames access only the motion vector data, and thus
avoid the huge memory traffic induced by executing the CNNs (SRAM spills).
Specifically, each I-frame incurs 646~MB memory traffic whereas E-frames require
only 22.8~MB.

Finally, the second to last column in \Fig{fig:yolo_ped_energy} shows the total
energy of EW-8 when extrapolation is performed on CPU. EW-8 with CPU-based
extrapolation consumes almost as high energy as EW-4, essentially negating the
benefits of extrapolation. This confirms that our architecture
choice of using a dedicated motion controller IP to achieve task
autonomy is important to realizing the full benefits in the vision pipeline.

\begin{figure*}[t]
\centering
\subfloat[\small{Success rate comparison.}]
{
  \includegraphics[trim=0 0 0 0, clip, height=1.6in]{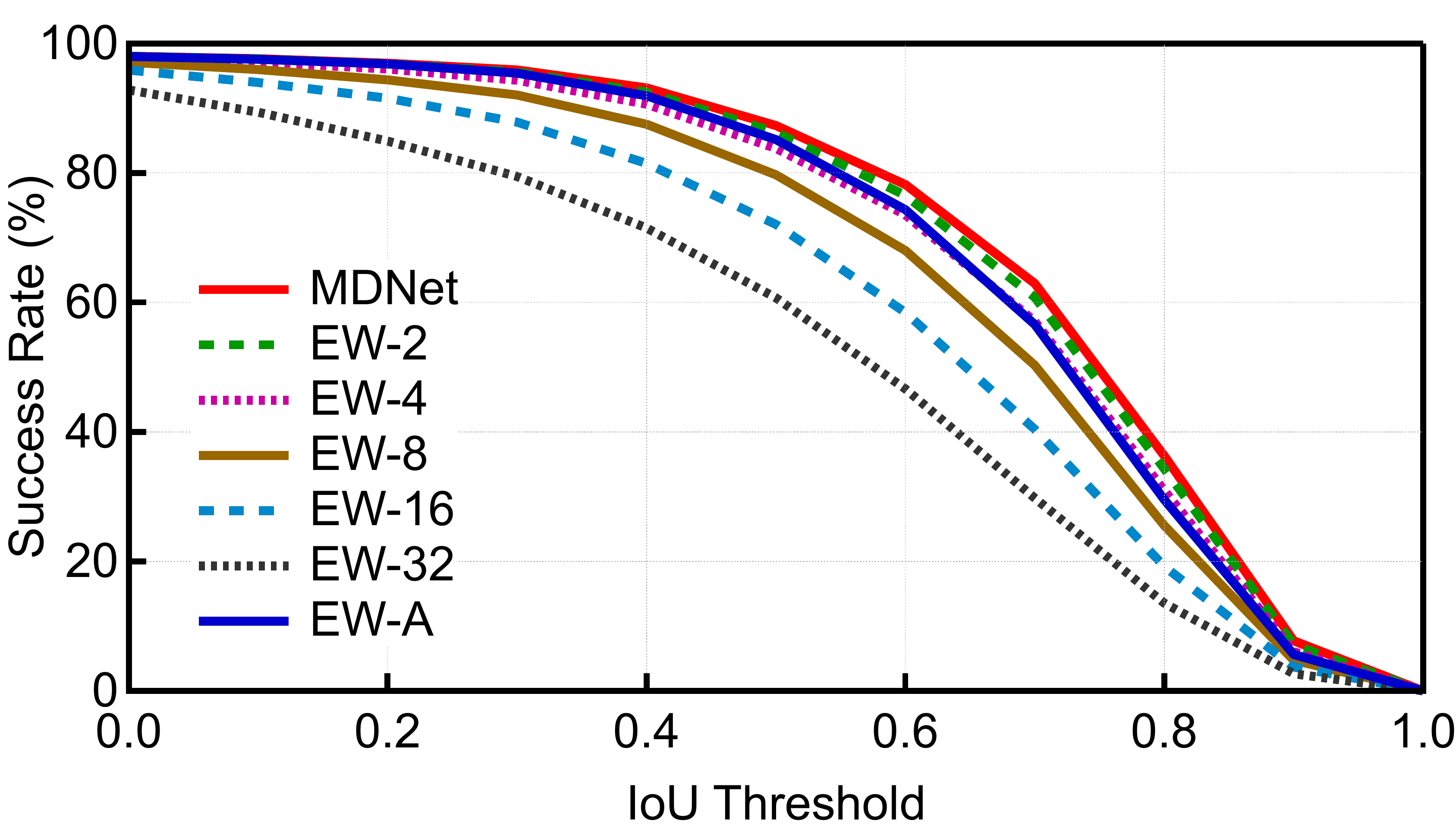}
  \label{fig:mdnet_accuracy}
}
\hspace*{10pt}
\subfloat[\small{Normalized energy consumption and inference rate comparison.}]
{
  \includegraphics[trim=0 0 0 0, clip, height=1.6in]{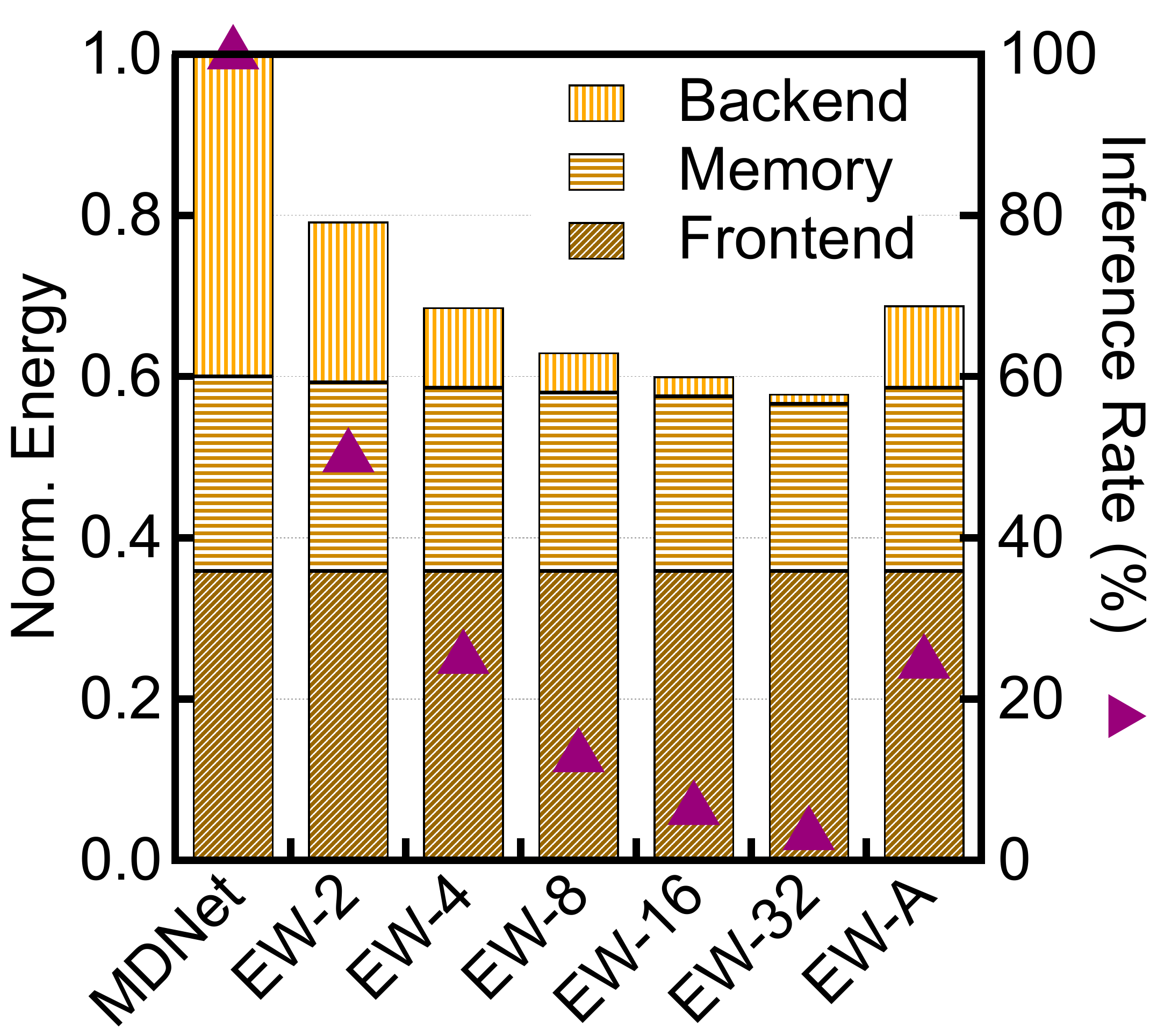}
  \label{fig:mdnet_energy}
}
\hspace*{10pt}
\subfloat[\small{Success rate for all 125 video sequences under an IoU ratio of 0.5.}]
{
  \includegraphics[trim=0 0 0 0, clip, height=1.6in]{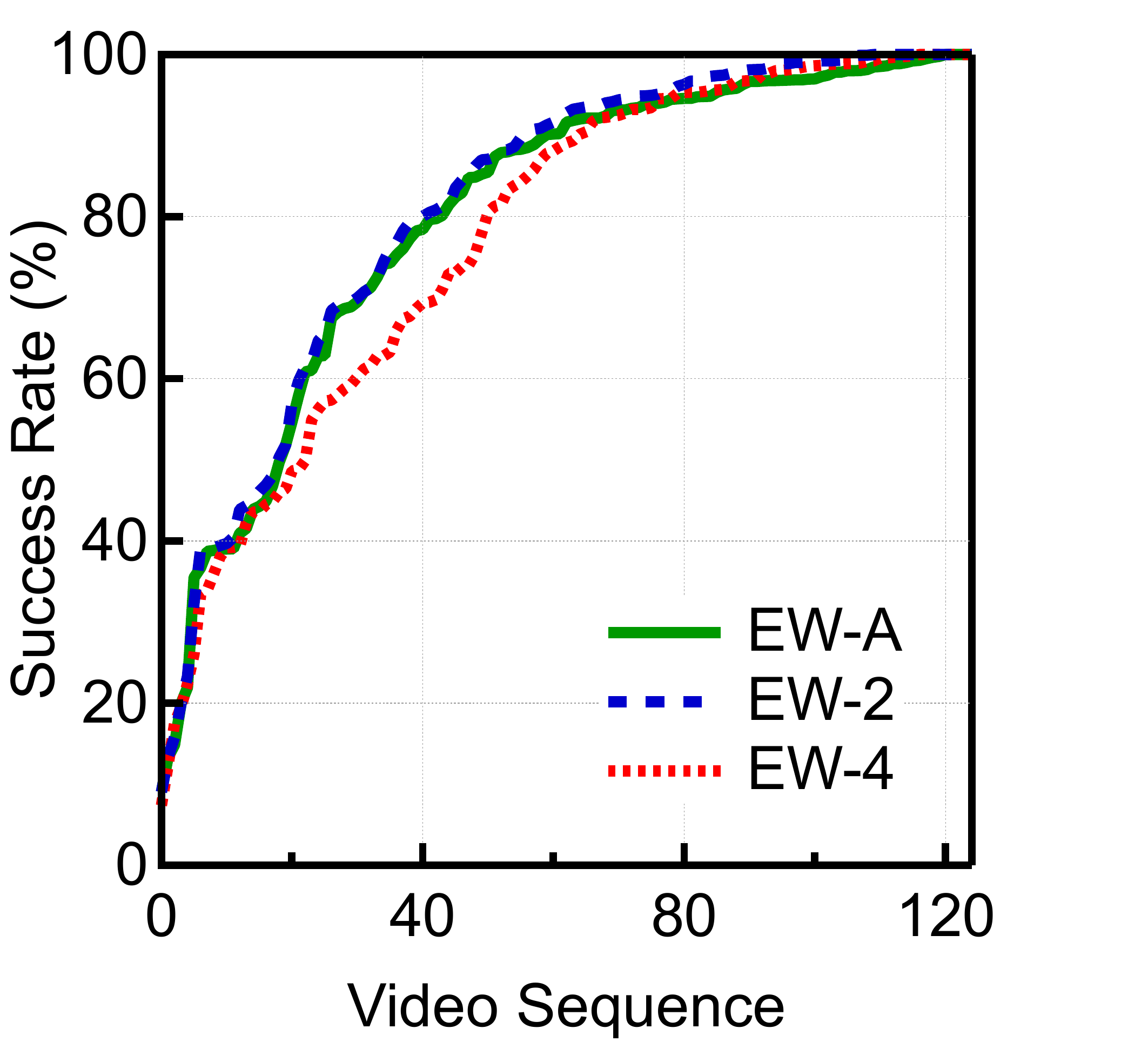}
  \label{fig:mdnet_accuracy_all}
}
\caption{\small Accuracy loss and energy saving comparisons between baseline MDNet and various Euphrates configurations for OTB 100 and VOT 2014 datasets. Energy is dissected into three parts: backend (CNN engine and motion controller), main memory, and frontend (sensor and ISP).}
\label{fig:mdnet_otb}
\end{figure*}

\paragraph{Tiny YOLO Comparison} One common way of reducing energy consumption and
improving FPS is to reduce the CNN model complexity. For instance, Tiny YOLO uses only nine of YOLOv2's 24 convolutional layers, and thus has an 80\% MAC operations reduction~(\Tbl{tab:workloads}).

However, we find that exploiting the temporal motion information is a more
effective approach to improve object detection efficiency than simply truncating a complex network. The bottom curve in \Fig{fig:yolo_ped_accuracy} shows the average
precision of Tiny YOLO. Although Tiny YOLO executes 20\% of YOLOv2's MAC
operations, its accuracy is even lower than EW-32, whose computation
requirement is only 3.2\% of YOLOv2. Meanwhile, Tiny YOLO consumes about 1.5
$\times$ energy at a lower FPS compared to EW-32 as shown in
\Fig{fig:yolo_ped_energy}.

\subsection{Visual Tracking Results}
\label{sec:eval:tracking}

Visual tracking is a simpler task than object detection. As shown
in~\Tbl{tab:workloads}, the baseline MDNet is able to achieve real-time (60 FPS)
frame rate using our CNN accelerator. This section shows that without degrading
the 60 FPS frame rate, Euphrates reduces energy consumption by 21\% for
the vision subsystem (50\% for the backend), with an 1\% accuracy loss.

\paragraph{Results} \Fig{fig:mdnet_accuracy} compares the success rate of
baseline MDNet, Euphrates under different EW sizes (EW-N, where N ranges from 2
to 32 in power of 2 steps), and Euphrates under the adaptive mode (EW-A).
Similar to object detection, reducing the CNN inference rate using motion
extrapolation incurs only a small accuracy loss for visual tracking. Specifically,
under an IoU of 0.5 EW-2's accuracy degrades by only 1\%.

The energy saving of Euphrates is notable. \Fig{fig:mdnet_energy} shows the
energy breakdown under various Euphrates configurations normalized to the
baseline MDNet. As a reference, the right $y$-axis shows the inference rate,
i.e., the percentage of frames where a CNN inference is triggered. EW-2 and EW-4
trigger CNN inference for 50\% and 25\% of the frames, thereby achieving
21\% and 31\% energy saving compared to MDNet that has an 100\%
inference rate. The energy savings are smaller than in the object detection
scenario. This is because MDNet is simpler than YOLOv2~(\Tbl{tab:workloads}) and
therefore, the vision backend contributes less to the overall energy
consumption.

Finally, Euphrates provides an energy-accuracy trade-off through
tuning EW.
For instance, EW-32
trades 42\% energy savings with about 27\% accuracy loss at 0.5 IoU.
Extrapolating further beyond 32 frames, however, would have only marginal energy
savings due to the increasing dominance of the vision frontend and
memory, as is evident in \Fig{fig:mdnet_energy}.

\paragraph{Adaptive Mode} Compared to the baseline MDNet, the adaptive mode of
Euphrates (EW-A) reduces energy by 31\% with a small accuracy loss of
2\%, similar to EW-4. However, we find that the adaptive mode has a more uniform
success rate across different tracking scenes compared to EW-4.
\Fig{fig:mdnet_accuracy_all} shows the success rate of all 125 video sequences
in the OTB 100 and VOT 2014 datasets under EW-A, EW-2, and EW-4, sorted from low
to high. Each video sequence presents a different scene that varies in terms of
tracked objects (e.g., person, car) and visual attributes (e.g., occlusion,
blur). EW-A has a higher success rate than EW-4 across most of the scenes,
indicating its wider applicability to different object detection scenarios.
Compared to EW-2, the adaptive mode has a similar accuracy behavior, but
consumes less energy.

\subsection{Motion Estimation Sensitivity Study}
\label{sec:eval:sen}

Euphrates leverages motion vectors generated by the ISP. This section shows that the results of Euphrates are robust against
different motion vector characteristics. In
particular, we focus on two key characteristics: \textit{granularity} and
\textit{quality}.
Due to the space
limit we only show the results of object tracking on the OTB 100 dataset, but
the conclusion holds generally.



\begin{figure}[t]
\centering
\subfloat[\small{Sensitivity of success rate to different macroblock sizes.}]
{
  \includegraphics[trim=0 0 0 0, clip, height=1.5in]{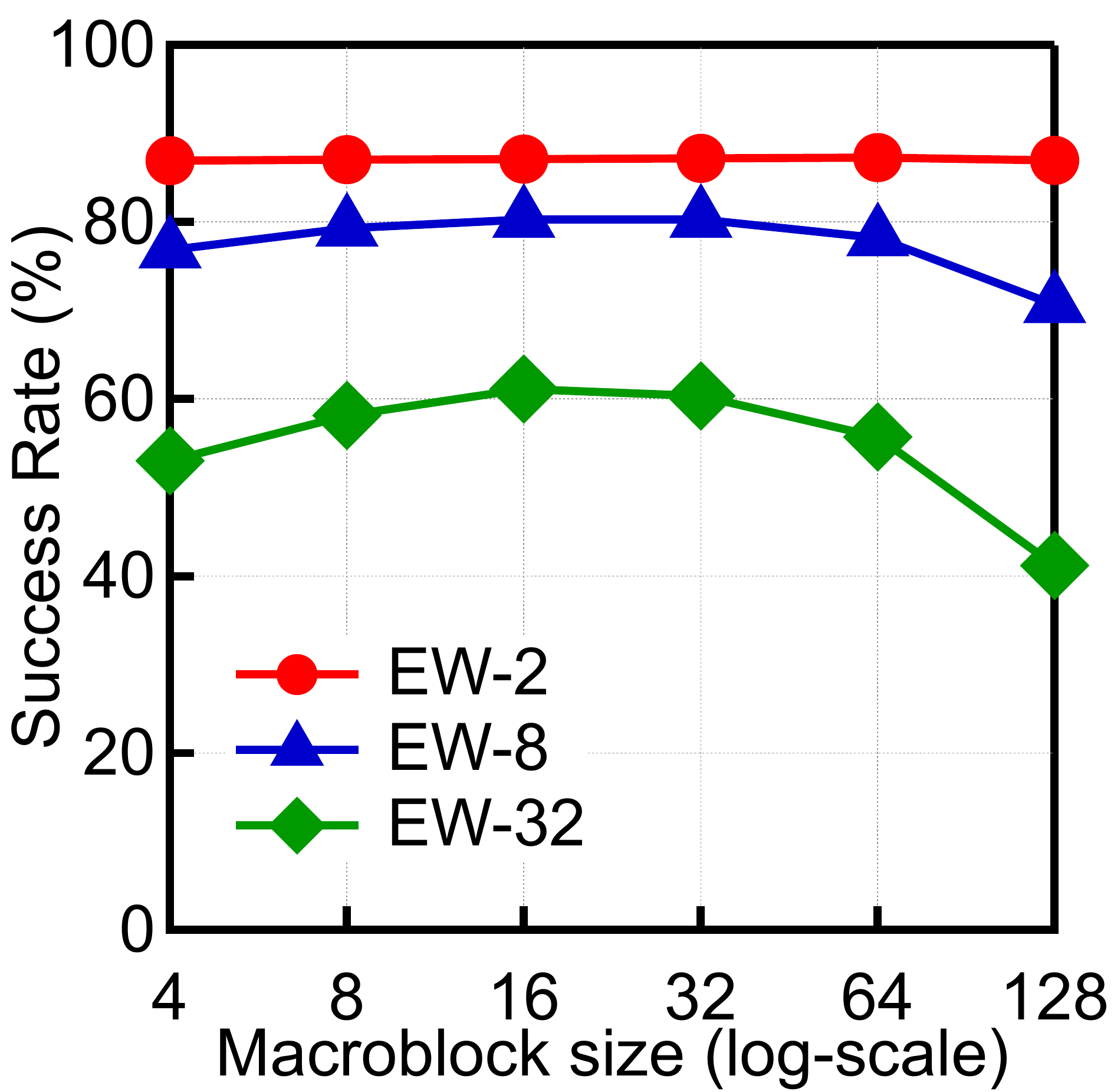}
  \label{fig:mbsize_all}
}\hfill
\subfloat[\small{Success rate comparison between ES and TSS.}]
{
  \includegraphics[trim=0 0 0 0, clip, height=1.5in]{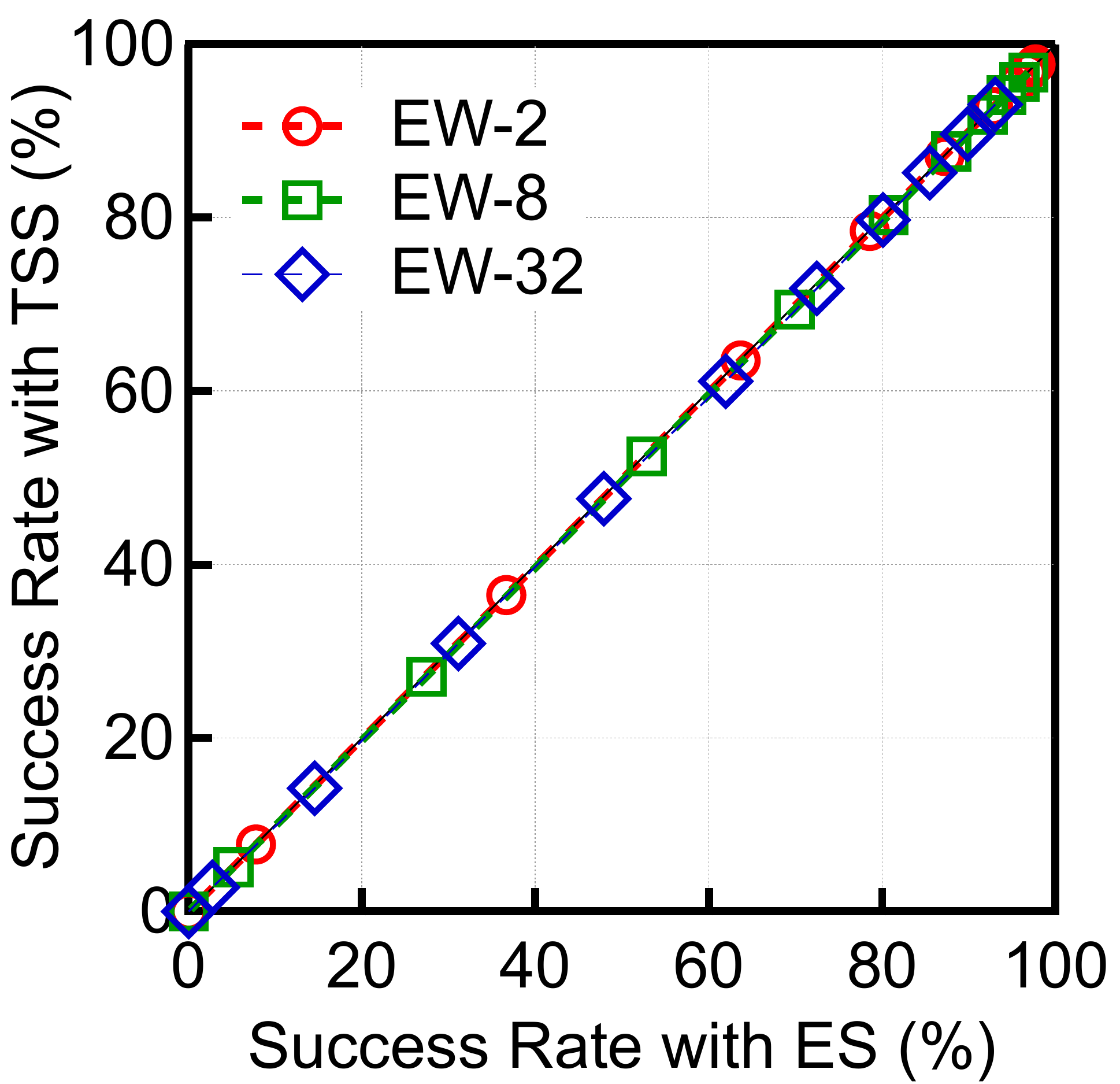}
  \label{fig:mv_accuracy}
}\hfill
\caption{\small Accuracy sensitivity to motion estimation results.}
\label{fig:mv_sen}
\end{figure}

\paragraph{Granularity Sensitivity} Motion vector granularity is captured by the
macroblock (MB) size during block-matching. Specifically, a $L \times L$ MB
represents the motions of all its $L^2$ pixels using one single vector.
\Fig{fig:mbsize_all} shows how the success rate under an IoU ratio of 0.5
changes with the MB size across three extrapolation windows (2, 8, and 32).


We make two observations. First, Euphrates' accuracy is largely insensitive to
the MB size when the extrapolation window is small. For instance, the
success rates across different MB sizes are almost identical in
EW-2. As the extrapolation window grows from 2 to 32, the impact of motion
vector granularity becomes more pronounced. This is because errors due to
imprecise motion estimation tend to accumulate across frames under a large
extrapolation window.



Second, MBs that are too small (e.g., 4) or too large (e.g., 128) have a
negative impact on accuracy~(\Fig{fig:mbsize_all}). This is because overly-small
MBs do not capture the global motion of an object,
especially objects that have deformable parts such as a running athlete;
overly-large MBs tend to mistake a still
background with the motion of a foreground object. $16\times16$ strikes the
balance and consistently achieves the highest success rate under large
extrapolation windows. It is thus a generally preferable motion vector granularity.

\paragraph{Quality Sensitivity} Motion vector quality can be captured by the Sum
of Absolute Differences (SAD) between the source and destination macroblocks.
Lower SAD indicates higher quality. Different block-matching algorithms
trade-off quality with computation cost. As discussed
in~\Sect{sec:background:mv}, the most accurate approach, exhaustive search (ES),
leads to $9\times$ more operations than the simple TSS algorithm~\cite{tss}.

We find that ES improves the accuracy only marginally over TSS. \Fig{fig:mv_accuracy} is a scatter plot comparing the success rates between ES
($x$-axis) and TSS ($y$-axis) where each marker corresponds to a different IoU
ratio. Across three different extrapolation window sizes (2, 8, and 32), the
success rates of ES and TSS are almost identical. 

\section{Discussion}
\label{sec:disc}

\begin{figure}[t]
  \centering
  \includegraphics[trim=0 0 0 0, clip, width=.9\columnwidth]{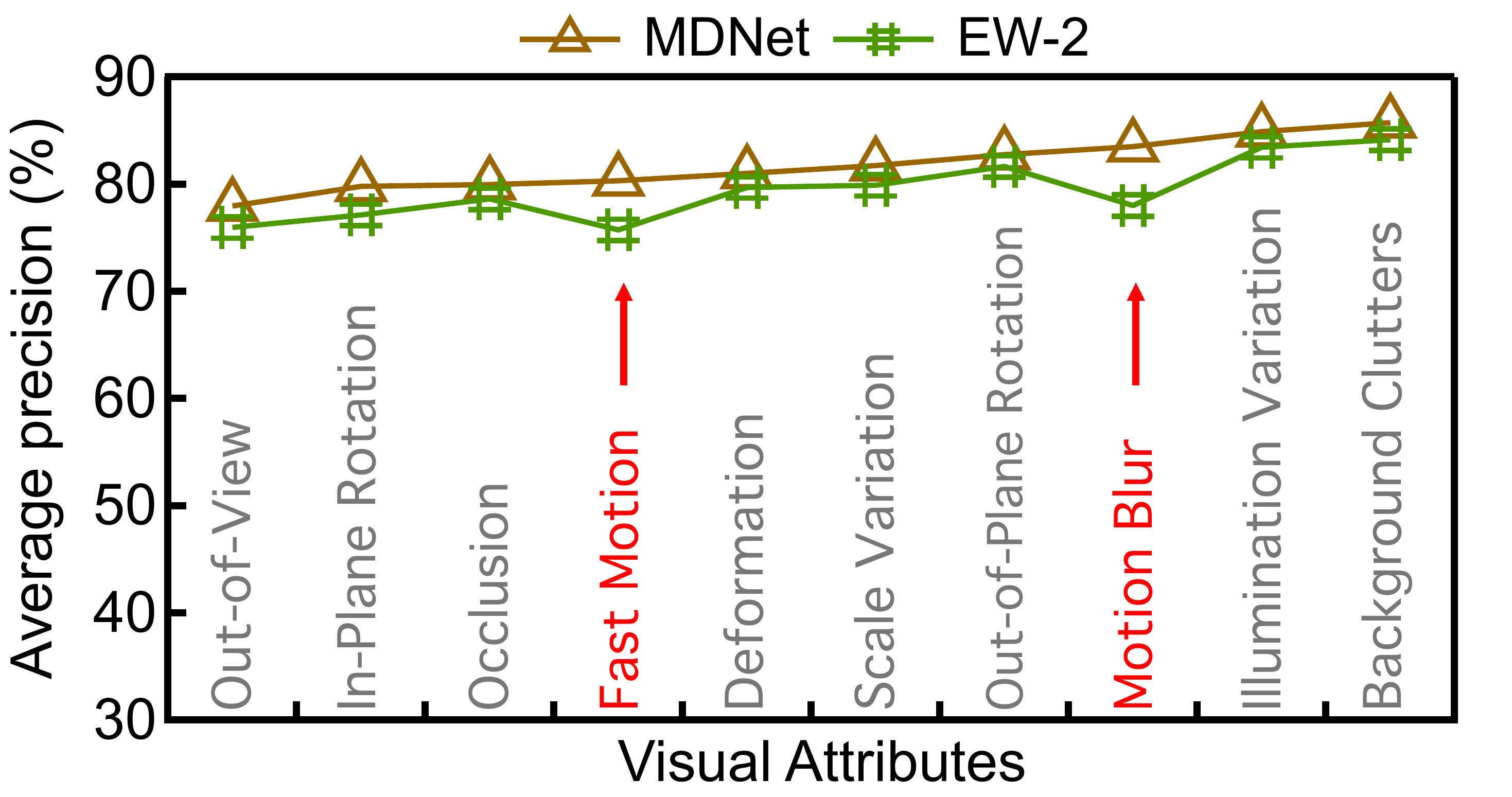}
  \caption{\small{Accuracy sensivitity to different visual attributes.}}
  \label{fig:mdnet_otb_attr}
\end{figure}

\paragraph{Motion Estimation Improvements} Euphrates is least
effective when dealing with scenes with fast moving and blurred objects. Let us
elaborate using the OTB 100 dataset, in which every video sequence is annotated
with a set of visual attributes~\cite{otbdataset} such as illumination
variation.  \Fig{fig:mdnet_otb_attr} compares the average precision between MDNet and EW-2 across
different visual attribute categories.
Euphrates introduces the most accuracy loss in the Fast Motion and Motion Blur
category.


Fast moving objects are challenging due to the limit of the search window size
of block matching (\Sect{sec:background:mv}), in which an accurate match is
fundamentally unobtainable if an object moves beyond the search window.
Enlarging the search window might improve the accuracy, but has significant
overhead. Blurring is also challenging because block-matching might return a
macroblock that is the best match for a blurred object but does not represent
the actual motion.

In the short term, we expect that fast and blurred motion can be greatly alleviated by higher frame
rate (e.g., 240 FPS) cameras that are already available on consumer mobile
devices~\cite{iphone6slowmo}. A high frame rate camera reduces the amount of
motion variance in consecutive frames and also has very short exposure time that
diminishes motion blur~\cite{nfs}. In the long term, we believe that it is critical to 
incorporate non-vision sensors such as an Inertial Measurement Unit~\cite{mpu9250}
as alternative sources for motion~\cite{applemotioncoproc, riftimu}, and combine vision vs. non-vision sensors for accurate motion estimation, as exemplified in the video stabilization feature in the Google Pixel 2 smartphone~\cite{pixel2motion}.


\paragraph{Hardware Design Alternatives} The video codec is known for using
block-matching algorithms for video
compression~\cite{videocodec}. However, video codecs are trigged only if
real-time camera captures are to be stored as video files. This, however, is not the
dominant use case in continuous vision where camera captures are consumed as
sensor inputs by the computer vision algorithms, rather than by humans.

That being said, video codecs complement ISPs in scenarios where ISPs are not
naturally a part of the image processing pipeline. For instance, if video frames
are streamed from the Internet, which are typically compressed, or are retrieved
from compressed video files, video codecs can readily provide motion vectors to
augment the continuous vision tasks whereas ISPs typically are idle during these
tasks.  We leave it as future work to co-design the codec hardware with the vision backend.






\section{Related Work}
\label{sec:related}

\paragraph{Motion-based Continuous Vision} Euphrates exploits the temporal
motion information for efficient continuous vision, an idea recently getting
noticed in the computer vision community. Zhang et al.~\cite{mvar} and Chadha et
al.~\cite{mvcnn} propose CNN models for action recognition using motion
vectors as training inputs. TSN~\cite{tsn} trains a CNN directly using multiple
frames. Our algorithm differs from all above in that Euphrates does not require
any extra training effort. In the case of TSN which is also evaluated using OTB
100 datasets, our algorithm ends up achieving about 0.2\% higher accuracy.


Fast YOLO~\cite{fastyolo} reuses detection results from the previous frame if
insufficient motion is predicted. Euphrates has two advantages. First, Fast YOLO
requires training a separate CNN for motion prediction while Euphrates leverages
motion vectors. Second, Fast YOLO does not perform extrapolation due to the lack
of motion estimation whereas Euphrates extrapolates using motion vectors to
retain high accuracy.

Finally, Euphrates is composable with all the above systems as they can be used as
the baseline inference engine in Euphrates. Our algorithm and hardware
extensions do not modify the baseline CNN model, but improves its
energy-efficiency.


\paragraph{Energy-Efficient Deep Learning} Much of the recent research on energy-efficient CNN architecture
has focused on designing better accelerators~\cite{tpu, eyeriss,
cnnresource, cnvlutin, shidiannao}, leveraging emerging memory
technologies~\cite{prime, neurocube, pipelayer}, exploiting sparsity and
compression~\cite{eie, circnn, scnn, scalpel, whatmough2017isscc}, and better tooling~\cite{minerva,
neutrams, tabla}. Euphrates takes a different but complementary approach. While
the goal of designing better CNN architectures is to reduce energy consumption per
inference, Euphrates reduces the rate of inference by replacing inferences with
simple extrapolation and thus saves energy.

Suleiman et al.~\cite{energygap} quantitatively show that classic CV algorithms
based on hand-crafted features are extremely energy-efficient at the cost of
large accuracy loss compared to CNNs. Euphrates shows that motion extrapolation
is a promising way to bridge the energy gap with little accuracy loss.


\paragraph{Specialized Imaging \& Vision Architectures} Traditional ISPs perform only primitive image processing while leaving advanced photography and vision tasks such as high dynamic range and motion estimation to CPUs and GPUs. Modern imaging and
vision processors are capable of performing advanced tasks \textit{in-situ} in
response to the increasing compute requirements of emerging algorithms. For
instance, Google's latest flagship smartphone, Pixel 2, has a dedicated SoC
equipped with eight Image Processing Units for advanced computational
photography algorithms such as HDR+~\cite{pixelsoc}. IDEAL~\cite{ideal}
accelerates the BM3D-based image denoising algorithms. Mazumdar et
al.~\cite{compcommcam} design specialized hardware for face-authentication and
stereoscopic video processing.


Researchers have also made significant effort to retain programmability in
specialized vision architectures~\cite{convengine, ivs, paisp}. For instance, Movidius Myriad
2~\cite{myriad2} is a VLIW-based vision processor used in Google Clip
camera~\cite{myriad24clip} and DJI Phantom 4 drone~\cite{myriad24phantom4}.
Clemons et al.~\cite{patchmem} propose a specialized memory system for imaging
and vision processors while exposing a flexible programming interface. Coupled
with the programmable architecture substrate, domain specific languages such as
Halide~\cite{halide}, Darkroom~\cite{darkroom}, and Rigel~\cite{rigel} offer
developers even higher degree of flexibility to implement new features. We expect imaging and vision processors in future to be highly programmable,
offering more opportunities to synergistically architect image processing and
continuous vision systems together as we showcased in this paper.


\paragraph{Computer Vision on Raw Sensor Data} Diamond et al.~\cite{rawsensor}
and Buckler et al.~\cite{ispconfig} both showed that CNN models can be
effectively trained using raw image sensor data. RedEye~\cite{redeye} and ASP
Vision~\cite{asp} both move early CNN layer(s) into the camera sensor and
compute using raw sensor data. This line of work is complementary to Euphrates
in that our algorithm makes no assumption about which image format motion
vectors are generated. In fact, recent work has shown that motion can be
directly estimated from raw image sensor data using block
matching~\cite{boracchi2008multiframe, yang2017evolutionary}. We leave it as
future work to port Euphrates to support raw data.

\section{Conclusion}
\label{sec:conc}

Delivering real-time continuous vision in an energy-efficient manner is a tall order for mobile system
design. To overcome the energy-efficiency barrier, we must expand the research
horizon from individual accelerators toward holistically co-designing different mobile SoC components. This paper demonstrates one such co-designed system to enable motion-based synthesis. It leverages the temporal motion information naturally produced by the imaging engine (ISP) to reduce the compute demand of the vision engine (CNN accelerator).


Looking forward, exploiting the synergies across different SoC IP blocks will become ever more important as mobile SoCs incorporate more specialized domain-specific IPs. Future developments should explore
cross-IP information sharing beyond just motion metadata and expand the co-design scope to other on/off-chip
components. Our work serves the first step, not the final word, in a promising new
direction of research.

\raggedright
\balance
\bibliographystyle{IEEEtranS}
\bibliography{refs}

\end{document}